\pdfoutput=1
\documentclass[11pt]{article}

\usepackage{emnlp2022}

\usepackage[utf8]{inputenc} % allow utf-8 input
\usepackage[T1]{fontenc}    % use 8-bit T1 fonts
\usepackage{hyperref}       % hyperlinks
\usepackage{url}            % simple URL typesetting
\usepackage{booktabs}       % professional-quality tables
\usepackage{amsfonts}       % blackboard math symbols
\usepackage{nicefrac}       % compact symbols for 1/2, etc.
\usepackage{microtype}      % microtypography
\usepackage{xcolor}         % colors
\usepackage{inconsolata}    % typewriter font aesthetics

\usepackage{times}
\usepackage{latexsym}
\usepackage{amsmath}
\usepackage{pifont}
\usepackage{graphicx}
\usepackage{listings}

\usepackage{subcaption}

\title{Sparse Mixers: Combining MoE and Mixing to build a more efficient BERT}

\author{
  James Lee-Thorp \and Joshua Ainslie \\
  Google Research \\
  \texttt{\{jamesleethorp, jainslie\}@google.com}
}

\begin{document}

\maketitle

\begin{abstract}
We combine the capacity of sparsely gated Mixture-of-Experts (MoE) with the speed and stability of linear, mixing transformations to design the Sparse Mixer encoder model. Sparse Mixer slightly \emph{outperforms} BERT on GLUE and SuperGLUE, but more importantly trains 65\% faster and runs inference 61\% faster. We also present a faster variant, prosaically named Fast Sparse Mixer, that marginally \emph{underperforms} BERT on SuperGLUE, but trains and runs nearly twice as fast. We justify the design of these two models by carefully ablating through various mixing mechanisms, MoE configurations, and hyperparameters. Sparse Mixer overcomes many of the latency and stability concerns of MoE models and offers the prospect of serving sparse student models, without resorting to distilling them to dense variants.\footnote{Source code available at \url{https://github.com/google-research/google-research/tree/master/sparse_mixers}.}
\end{abstract}

\section{Introduction}
\label{sec:introduction}

Sparsely gated Mixture-of-Experts (MoE) models have seen a surge of interest in recent years \citep{shazeer2017outrageously, lepikhin2020gshard, fedus2021switch, riquelme2021scaling, du2021glam, artetxe2021efficient, clark2022unified, mustafa2022multimodal}. MoE models offer the promise of sublinear compute costs with respect to the number of model parameters. By training "experts" that can independently process different slices of input data, MoE layers increase model capacity with limited increases in FLOPS. 

Perhaps because of the favorable capacity-to-compute trade-off, most recent MoE studies, including the aforementioned, have focused on using MoE to scale up large models. Using MoE layers to scale to larger models offers quality and total train efficiency gains over dense models, but not train or inference step latency benefits.
Indeed, the task of serving these models in practice is either ignored or relegated to distilling the sparse teacher model to a dense student model \citep{hinton2015distilling}, often with a significant quality loss relative to the sparse teacher model. For example, \citet{fedus2021switch} are only able to distill roughly $30\%$ of the Switch Transformer's quality gains to a dense model.

Orthogonal to MoE, efficient mixing models \citep{tolstikhin2021mlp, liu2021pay, lee2021fnet} replace attention in Transformer-like models with simpler linear transformations or MLP blocks that "mix" input representations. Linear transformations are particularly attractive because they are faster than the combined projection and dot product operations in an attention layer.

In this work, we pull on both MoE and mixing threads to build low latency, sparse encoder models that we hope can used in production settings. We focus on encoder models, and BERT-like models in particular, because they are widely used in practice -- for example, in dual encoders for retrieval \citep{bromley1993signature, karpukhin2020dense}. 
% In future work, we plan to investigate porting the recipes outlined in this paper to encoder-decoder models \citep{raffel2019exploring}.

Relative to the vanilla Transformer model \citep{vaswani2017attention}, we speed up our model in two ways. (1) We use the increased capacity from MoE sublayers to offset parameter reductions in other parts of the model. (2) We use mixing transformations to replace a large fraction of self-attention sublayers with faster, linear transformations. The resulting model, which we name Sparse Mixer, slightly ($<1\%$) \emph{outperforms} BERT on GLUE \citep{wang2018glue} and SuperGLUE \citep{wang2019superglue}, but most importantly trains 65\% faster and runs inference 61\% faster. We also introduce a simple variant of Sparse Mixer, prosaically named Fast Sparse Mixer, that marginally ($<0.2\%$) \emph{under-performs} BERT on SuperGLUE, but runs nearly twice as fast: training 89\% faster and running inference 98\% faster.

An interesting finding of our work is a training stability synergy between the sparse and mixing model components.
% Indeed, MoE models are often prone to instability during training \citep{zoph2022designing}.
As a point of comparison, we find that simply replacing dense feed-forward sublayers in BERT with MoE variants yields highly unstable models; see Section \ref{sec:sparse_mixer}. However, these instabilities dissipate as we replace self-attention sublayers with mixing sublayers. We hypothesize that the (token-dependent) relevance weighted self-attention basis is the source of the instability, and hence that replacing the majority of self-attention sublayers with mixing sublayers renders sparse mixer models highly stable.

In summary, we introduce two models:
\begin{itemize}
    \item Sparse Mixer, which matches BERT on GLUE and SuperGLUE but runs 61-65\% faster.
    \item Fast Sparse Mixer, which slightly underperforms BERT ($<$0.2\%) but is nearly $2$x faster.
\end{itemize}
We justify the design of these models by ablating through model mixing, MoE, and hyperparameter configurations. With Sparse Mixers, we demonstrate that the speed and stability regressions of MoE models may be overcome using mixing mechanisms. This offers the promise of directly serving sparse models, rather than resorting to distilling them to dense variants.

\section{Related work}
\label{sec:related_work}

% \subsection{Sparsely gated Mixture-of-Experts}
% \label{subsec:moe}

\textbf{Sparsely gated Mixture-of-Experts.} Mixture-of-Experts (MoE) models were introduced by \citet{jacobs1991adaptive, jordan1994hierarchical} and more recently popularized by \citet{shazeer2017outrageously}. Recent work, such as \citep{zoph2022designing}, has played out the promise of MoE models by achieving state of the art results on a number of NLP benchmarks. As with contemporary MoE studies \citep{du2021glam, lepikhin2020gshard}, these models are large and primarily focus on model quality. When efficiency is studied, it is typically at the level of a total train time efficiency metric. For example, although the per training step speed of the Switch Transformer \citep{fedus2021switch} is slower than the vanilla Transformer, because the Switch Transformer surpasses the vanilla model's top accuracy in a fraction of the steps, the Switch Transformer can be correctly described as a more efficient model. However, the slower step speed is an Achilles heel for serving such models; 
% for low latency settings, 
one generally cannot ask a user to wait longer for a more accurate model response.

A notable exception is \citet{jaszczur2021sparse}, who sparsify multiple components of the Transformer, primarily by replacing softmaxes with argmaxes, to achieve an over 2x speed-up in \emph{unbatched} inference speed on CPUs for Base/Large model sizes
% , with larger speed-ups for very large models and for long sequence inputs.
In contrast to our work, their speed-ups to not carry over to accelerator hardware or to training. 
% MoE algorithms, in contrast, offer a more hardware accelerator friendly sparse algorithm as each expert perform dense matrix multiplications.
%
% In this work, we focus on \emph{per step latency} for both training and inference on GPUs and TPUs.

Memory mechanisms are another popular sparse technique for adding capacity to models with limited increases in compute; see, for example, \citep{weston2015memory, sukhbaatar2015end, lample2019large}. While intuitively appealing and empirically promising, suboptimal implementations (look-ups in particular) for accelerator hardware often yield memory models that have favorable theoretical compute properties, but are slow in practice.

% \subsection{Mixing}
% \label{subsec:mixing}

\textbf{Mixing.} Several recent works have explored mixing mechanisms, such as matrix multiplications \citep{tay2020synthesizer, lee2021fnet}, MLP blocks \citep{tolstikhin2021mlp, liu2021pay}, and spectral transforms \citep{lee2021fnet}, as an efficient replacement of attention in Transformer-like models. \citet{you2020hard, raganato2020fixed, lee2021fnet} find that hybrid attention-mixing models, wherein partial or a limited number of attention sublayers are retained, were faster than Transformers with only very limited accuracy degradation. Building off these works, we use sparse MLP sublayers to compensate for the remaining accuracy gap.

% \subsection{Model parameter configuration}
% \label{subsec:model_shape}

\textbf{Model parameter configuration.} Scaling up models has proven to be a successful program for increasing model quality \citep{kaplan2020scaling, raffel2019exploring, brown2020language}. The relationship between the number of model parameters and model quality can be roughly modelled through a power law \citep{kaplan2020scaling, clark2022unified, hoffmann2022training}. However, the configuration of these parameters within the model also plays an important role in model quality and efficiency. Consistent with \citet{tay2021scale}, we find that making the model thinner (smaller model dimensions) but deeper (more layers) is generally an efficient way to distribute parameters throughout the model.

\textbf{Distillation.} Knowledge distillation \citep{hinton2015distilling} is a powerful technique that has been successfully deployed to train efficient "student" BERT models from larger "teacher" models \citep{sanh2019distilbert, jiao2019tinybert, sun2019patient, xu2020bert}. Although we suspect that Sparse Mixer will offer a promising distillation architecture, we view the distillation techniques themselves as orthogonal to our architecture optimization goal. Indeed, we show, in Figure \ref{fig:mlm_scale}, that speed-ups and quality gains from Sparse Mixer carry over to both larger (teacher) and smaller (student) sizes.

\section{Model}
\label{sec:model}

\subsection{Architecture}
\label{subsec:architecture}

% \begin{figure}
%     \centering
%     \includegraphics[width=0.45\textwidth]{figures/general_architecture.pdf}
%     \caption{Block based encoder architecture. The model has $N$ encoder blocks, each containing mixing and MLP sublayers. Each MLP sublayer may be sparse or dense. Each mixing sublayer may use self-attention or a mixing transformation.}
%     \label{fig:general_architecture}
% \end{figure}

\begin{figure}[tb]
    \centering
    \includegraphics[width=0.45\textwidth]{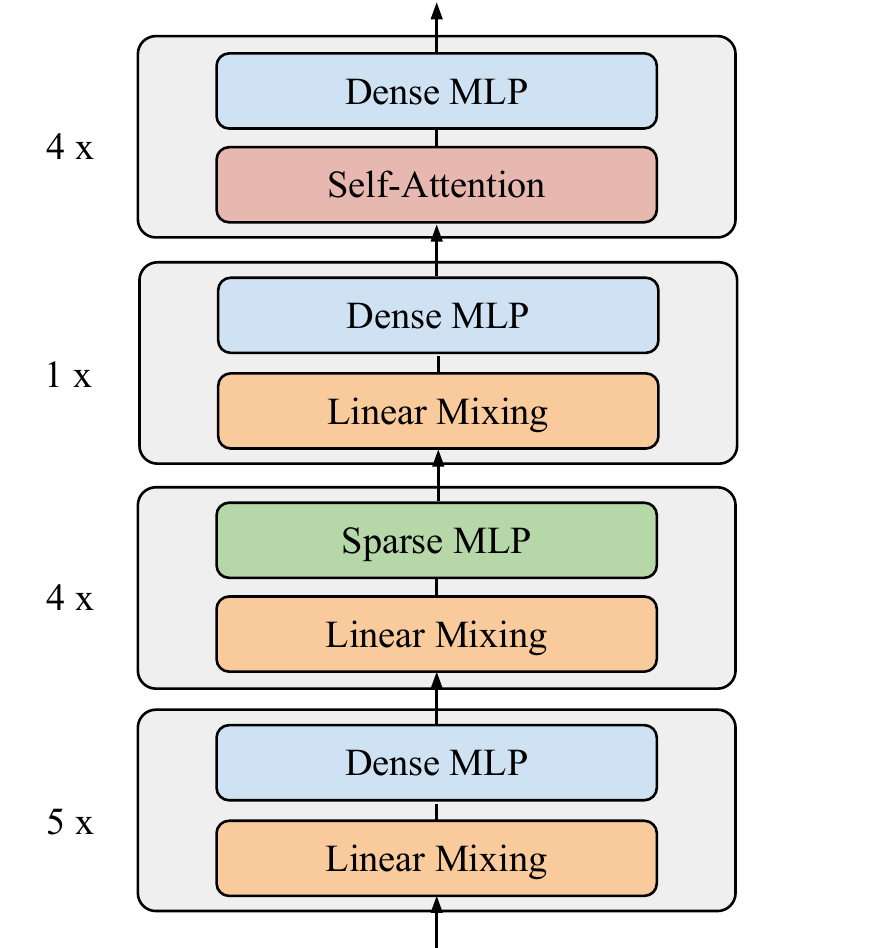}
    \caption{Sparse Mixer encoder blocks for the Base configuration. Layer norms, residual connections, embedding layers and output layers are not shown. The top $K=4$ blocks contain self-attention and dense MLPs; the middle $M=4$ blocks contain mixing and sparse MLPs; and the remaining $L=1$ and $P=5$ blocks contain mixing and dense MLPs.}
    \label{fig:sparse_mixer_architecture}
\end{figure}

Our design space for the Sparse Mixer builds off of the stacked encoder blocks of BERT \citep{devlin2018bert}, which we use as our canonical Transformer encoder \citep{vaswani2017attention}. Each encoder block contains a mixing or self-attention sublayer and a (dense or MoE) MLP sublayer, connected with residual connections and layer norms. We keep the standard BERT input embedding and output projection layers \citep{devlin2018bert}; see also Appendix \ref{subsec:bert_arch}. We arrive at the Sparse Mixer encoder block stack, shown in Figure \ref{fig:sparse_mixer_architecture}, by carefully ablating through mixing mechanisms, MoE configurations, and model hyperparameters in Section \ref{sec:coord_descent}.

\subsection{MoE}
\label{subsec:moe_background}

% % When training a dense model layer over multiple devices, the first tool we reach for is data parallelism: we replicate our layer over all devices and perform the same model computations in parallel over separate shards of the data. In contrast, when training an MoE layer over multiple devices, we use data and expert parallelism: we initialize multiple, different instances ("experts") of our layer over the devices and perform parallel computations with each instance over separate shards of the data. 

% % Importantly, there is a trade-off between these two setups: the MoE layer has more capacity than the dense layer but each parameter receives less training signal. Indeed, the number of model parameters scales proportionally with the number of experts, but the gradient update for an individual sparse parameter is accumulated over less data relative to the replicated parameters of the dense layer.\footnote{In this work, we ignore model parallelism, an important technique for partitioning large models over multiples devices.}

In an MoE layer, we initialize multiple, different instances ("experts") of the layer and perform parallel computations with each instance over separate data shards. The sparsely activated MoE layers therefore have greater capacity than dense layers.
% % The number of experts in a given sparse MoE layer may be greater than the number of devices. 
As we increase the number of experts, we typically decrease the \emph{expert capacity} -- the number of tokens processed by an individual expert. More specifically, with $E$ denoting the number of experts and $n$ the number of tokens, we set 
\begin{equation*}
    % \label{eq:expert_capacity}
    \text{expert capacity} = cf \times n/E,
\end{equation*} where $cf$ is the scalar \emph{capacity factor}. For $cf \approx 1$, this allows us to increase model parameter count with minimal increases in FLOPS.\footnote{Modulo the typically relatively small increase in FLOPS from the token router.}
% ; see Section \ref{subsec:moe_background}.} 

% \subsubsection{Routing}
% \label{subsec:routing}

\textbf{Routing.} We use a \emph{router} or \emph{gating function} to carefully direct data shards between experts. This follows the intuition that expert \texttt{A} may become specialized at processing inputs in one part of the embedding space, while experts \texttt{B},  \texttt{C}, \ldots are specialized to other parts of the embedding space. 
It is the router that ensures sparsity by assigning only a subset of tokens to each expert, thereby ensuring that only a subset of parameters are activated for each token.

Router design is an active research area \citep{lewis2021base, roller2021hash, zhou2022mixture, clark2022unified}. We limit ourselves to two router types: traditional "Tokens Choose" and "Experts Choose". We follow the standard practice of routing at the token level -- the router assigns each token to a subset of experts. Both assignment algorithms first generate router logits by projecting token representations from the embedding dimension, $d_m$, to the expert dimension, $E$. 
% \begin{equation*}
%       \mathcal{W}: \mathbb{R}^{d_m} \mapsto \mathbb{R}^{E}.
% \end{equation*}
We apply a softmax to normalize the logits to a probability distribution. Finally, tokens are assigned to experts using one of the assignment algorithms.
% % \footnote{If the router logits blow up, the MoE model will become unstable. To encourage the model to produce small router logits, we use the router z-loss introduced by \citet{zoph2022designing}. 
% % We include the router z-loss in our study, with the recommended $10^{-4}$ scaling factor, although we do not observe instability issues without the router z-loss.
% % }

% \subsubsection{Tokens Choose routing}
% \label{subsec:tokens_choose}

\textbf{Tokens Choose routing.} For Tokens Choose routing \cite{shazeer2017outrageously}, each token is assigned to its top-k experts. We focus on top-1 ("Switch") routing \citep{fedus2021switch}. Because expert capacities are limited, there is no guarantee that a given token can be routed to its top expert, although any token that fails to reach an expert will still propagate into the next encoder block through the residual connection. There is also no guarantee that a given expert receives at least one token. So, to ensure that compute is efficiently distributed among experts, we include a load balancing loss as in \citep{shazeer2017outrageously, fedus2021switch}.

We can increase expert capacity by increasing the capacity factor, $cf$. This will increase the probability that a given token is routed to its desired experts. Decreasing $cf$ will further sparsify the model and speed up the MoE sublayer.\footnote{We can introduce separate train and evaluation capacity factors. Setting $cf_{\mathrm{eval}} > cf_{\mathrm{train}}$ may improve inference quality without slowing training, but will hurt inference speed.} We use Batch Prioritized Routing \citep{riquelme2021scaling} to prioritize routing tokens with the highest router probability, rather than simply routing tokens in the left-to-right ordering in the batch.

% \subsubsection{Experts Choose routing}
% \label{subsec:experts_choose}

\textbf{Experts Choose routing.} For the Experts Choose assignment algorithm \citep{zhou2022mixture}, experts choose their top tokens, rather than tokens choosing experts. This effectively amounts to a transpose of the router probabilities prior to the top-k operation. Each expert performs its top-k operation with k = \emph{expert capacity}. An individual token may be processed by multiple experts or none at all. Because experts have their choice of tokens and always fill their buffer, increasing the capacity factor, $cf$, will increase both the number of tokens that an expert processes and also the number of experts to which a given token is routed. Because each expert always fills its capacity, no auxiliary loading balancing loss is required.

% \subsubsection{Token group size}
% \label{subsec:group_size}

\textbf{Token group size.} Tokens are subdivided into groups and expert assignment is performed on a per-group basis. A larger group size will result in slower but more accurate top-k and sorting computations, whereas a smaller group size will result in faster but more approximate routing choices. In practice, we find that imperfect routing choices are tolerable and default to a group size of $4096$ tokens.

\textbf{Parallelization strategies}. In this work, we  focus on faster, servable architectures using expert and data parallelism. We use data parallelism to shard data across devices, and expert parallelism to partition experts across devices; for example, placing experts 1 and 2 on device 1, experts 3 and 4 on device 2, and so on. Model parallelism is a third axis to shard model weights (matrices) across devices; for example, expert 1 is split across devices 1 and 2, expert 2 is split across devices 3 and 4, and so on. Model parallelism is typically most beneficial for scaling to larger model sizes. 
% It can offer overall training time gains by virtue of the larger model asymptoting to a higher accuracy, but it does not in itself offer lower latency.
% Indeed, given a fixed batch size and number of devices, you typically get greater throughput by increasing data parallelism rather than model parallelism. 

\subsection{Mixing}
\label{subsec:mixing_background}

We use simple linear, mixing transformations as drop-in replacements for a subset of the self-attention sublayers. Mixing transformations offer speed for reduced capacity and flexibility. Indeed, the attention mechanism contains four parameterized projections and two dot product operations ("$QK$" and "$V$"), allowing self-attention sublayers to construct representations in a highly expressive, token-dependent basis. On the other hand, the mixing transformations that we investigate are implemented through two, token-independent projections, one along each of the sequence and model dimensions. Fixing the mixing basis, relative to different data inputs, turns out to stabilize the model.
% during training. 
% That said, in Section \ref{subsec:mixing_results}, we find that the most Pareto efficient models are hybrid models that include both mixing and some self-attention sublayers. 

% \subsubsection{Spectral transformations}
% \label{subsec:spectral}

\textbf{Spectral transformations.} We experiment with the Fourier and Hartley transforms
% , which showed the most promise of the spectral transforms studied by 
\citep{lee2021fnet}. We integrate these transforms through a Fourier sublayer. The Fourier sublayer applies a 1D Discrete Fourier Transform (DFT) along the sequence dimension, $\mathcal{F}_{\textrm{seq}}$, and a 1D DFT along the hidden dimension, $\mathcal{F}_{\textrm{h}}$:
\begin{equation}
    \label{eq:fourier_layer}
    y = \Re\left(\mathcal{F}_{\textrm{seq}}\left(\mathcal{F}_{\textrm{h}} (x)\right) \right) ,
\end{equation}
where $\Re$ denotes the real part.
% Following \citep{lee2021fnet}, we only keep the real part of the result.

The Hartley sublayer uses Equation \eqref{eq:fourier_layer} with the DFT replaced with the Discrete Hartley Transform, $\mathcal{H}$.\footnote{The Hartley Transform, which transforms real input to real output, can be described in terms of the Fourier Transform: $\mathcal{H} = \Re\left\{\mathcal{F}\right\} - \Im\left\{\mathcal{F}\right\}$. In the case of the Hartley Transform, we may omit the $\Re$ from Equation \eqref{eq:fourier_layer}.} We compute the Fourier and Hartley transforms using the Fast Fourier Transform (FFT) \citep{cooley1965algorithm, frigo2005design}.

In Equation \eqref{eq:fourier_layer}, we transform along both the sequence and hidden dimensions. Although the primary purpose of a mixing sublayer is to combine inputs along the sequence dimension, \citet{lee2021fnet} found that also mixing along the hidden dimension improved model quality.

% \subsubsection{Structured matrix projections}
% \label{subsec:structured}

\textbf{Structured matrix projections.} We explore structured matrices under the hypothesis that adding structure to the mixing basis may improve the distribution of output representations. We consider two parameterized, structured matrices: Toeplitz and circulant. A Toeplitz matrix is a matrix in which each diagonal is constant. A circulant matrix is a particular kind of Toeplitz matrix, in which all rows are composed of the same elements but rotated one element to the right relative to the preceding row. For both matrices, the weights are learned. The corresponding mixing sublayer mixes along the sequence and hidden dimension. For example, for the Toeplitz case, we perform:
\begin{equation}
    \label{eq:toeplitz_layer}
    y = \mathcal{T}_{\textrm{seq}} \ \mathcal{T}_{\textrm{h}} \ x ,
\end{equation}
where $\mathcal{T}_{\textrm{seq}}$ and $\mathcal{T}_{\textrm{h}}$ denote Toeplitz matrices.\footnote{Matrix multiplications involving circulant matrices can be computed using the FFT \citep{davis1970circulant}. 
This requires three operations: one FFT to diagonalize the computation, one to apply the diagonalized matrix multiplcation and one iFFT to transform back to real space. 
A Toeplitz matrix may be embedded in a circulant matrix to take advantage of the same FFT computation. In practice, we find that for standard sequence lengths (512), using the FFT is slower than direct matrix multiplications on both GPU and TPU.}

% \subsubsection{Vanilla matrix projections}
% \label{subsec:linear}

\textbf{Vanilla matrix projections.} We also consider "unstructured", fully dense parameterized matrix projections. Following \citep{lee2021fnet}, we call the mixing sublayer arising from this case, the "Linear" sublayer. The Linear sublayer performs the same FLOPS as the structured matrix sublayers (provided the FFT is not used), but is more flexible due to the increased number of matrix weights. 

\subsection{Implementation}
\label{subsec:implementation}

We train and optimize our model on 8 V100 GPUs. 
% Our ablations may yield slightly different speed-accuracy trade-offs on other accelerators (e.g. TPU), but 
We believe that our results are reasonably robust to differing accelerators (e.g. TPU) as almost all of our modifications boil down to accelerator friendly matrix multiplications. In Section \ref{sec:sparse_mixer}, we scale our model sizes up and down on TPUs and find that the same favorable efficiency trade-offs persist.
We use JAX \citep{jax2018github} in the Flax framework \citep{flax2020github}.\footnote{Sparse Mixers code is available at \url{https://github.com/google-research/google-research/tree/master/sparse_mixers}.}

\section{Coordinate Descent}
\label{sec:coord_descent}

We train in a typical transfer learning setting \citep{devlin2018bert}: Masked Language Modelling (MLM) and Next Sentence Prediction (NSP) pre-training, followed by fine-tuning on GLUE \citep{wang2018glue} and SuperGLUE \citep{wang2019superglue}. 
When comparing models, we always use the exact same setup for all models and baselines. In particular, we follow the pre-training setup in \citep{devlin2018bert} with a few updates: (1) we pre-train on the much larger C4 dataset \citep{raffel2019exploring}; (2) we use a $32000$ SentencePiece vocabulary model \citep{kudo2018sentencepiece} trained on a $100$ million sentence subset of C4; and (3) we use a smaller batch size of $64$ (\citet{devlin2018bert} uses $256$). We use a sequence length of $512$ throughout pre-training. Experiments are run on $8$ V100 GPU chips, except for the scaling experiments (Section \ref{sec:sparse_mixer}) which are run on 32 TPU v3 chips.

In this section, we follow a "coordinate descent" through our model configurations until we arrive at the final Sparse Mixer design. Given the large number of model hyperparameters to explore, we perform multiple parameter searches in parallel. For example, the model shape and MoE configurations are explored independently and then the most promising configurations from each program are combined. 
% Because we are optimizing one model configuration coordinate at a time,
% % -- some of which we perform in parallel -- 
% our manual gradient descent is pedagogical but potentially sub-optimal. It would be exciting to see future work expand both the coordinate space and jointly optimize multiple coordinates using Automated Machine Learning (AutoML) \citep{thornton2013auto, liu2018darts, peng2020pyglove}.

\begin{table}
    \caption{Average accuracy metrics and median pre-training step speeds for mixing models. The "Fourier" model is identical to FNet \citep{lee2021fnet}. Speed-ups relative to BERT (see Table \ref{tab:vitals}) are shown in parentheses. The best metrics are highlighted in boldface, while the second best metrics are underlined. Stars indicate the selected configurations.}
    \label{tab:pure_mixing}
    \centering
    \setlength{\tabcolsep}{5pt}
    \begin{tabular}{l | c  c  c | c}
        \hline
          & \multicolumn{3}{c|}{Accuracy (\%)} & Speed \\ 
         Model & GLUE & MLM & NSP & (ms/batch) \\ \hline \hline
        %  BERT & \textbf{82.1} & \textbf{63.8} & \textbf{81.8} & 300 \\ \hline
         Fourier & \textbf{78.4}  & 55.7 & 75.4 & \textbf{173 (1.75x)} \\
         Hartley $\star$ & \underline{78.0} & \textbf{58.5} & 74.9 & \textbf{172 (1.77x)} \\
         Circulant & 75.1 & \underline{58.3} & 75.6 & \underline{200 (1.52x)} \\
         Toeplitz & 76.5 & 57.7 & \underline{76.5} & \underline{200 (1.52x)} \\
         Linear $\star$ & 77.7 & 57.6 & \textbf{77.4} & \underline{200 (1.52x)} \\ \hline
    \end{tabular} 
\end{table}

For our coordinate descent study, we only pre-train for $500k$ steps, which we found to be reasonably indicative of model performance. Models are fine-tuned with the same batch size ($64$) on the Validation split of each respective GLUE task for $5$ epochs and the best result for each task is selected from across three default base learning rates, adapted from \citet{devlin2018bert}: $\{10^{-5}, 5 \cdot 10^{-5}, 10^{-4}\}$. Our final model is pre-trained for longer and is evaluated on both GLUE and SuperGLUE for a broader set of training configurations in Section \ref{sec:sparse_mixer}. 
% Given the reduced training setup, we only view the GLUE results in our coordinate descent study as indicative of relative model improvements.

We prioritize efficiency -- speed and accuracy. We use pre-training step speed as a proxy for model latency. We rely on downstream average GLUE scores as our primary accuracy metric, but fallback to upstream MLM and NSP accuracies when GLUE scores between model variants are similar. Additional coordinate descent experiments are summarized in Appendix \ref{subsec:extra_coord_descent}, and full GLUE results for all coordinate descent experiments are provided in Appendix \ref{subsec:full_glue}.

\subsection{Mixing}
\label{subsec:mixing_results}

\textbf{Mixing mechanisms.} We compare the mixing mechanisms discussed in Section \ref{sec:related_work}. For each mixing model, we first replace \emph{all} self-attention sublayers with the corresponding mixing sublayer. The results are shown in Table \ref{tab:pure_mixing}. The spectral models (Fourier and Hartley) perform the best on GLUE. The Linear model slightly under-performs the spectral models, while the structured mixing models (Circulant and Toeplitz) perform worst on GLUE. 
% The pre-training metrics are a little more muddled, with the Hartley and Circulant models performing relatively well on MLM and the Linear model performing best on NSP. 
The spectral methods, efficiently implemented using FFTs, are the fastest.\footnote{There is some noise in the speed measurement, so we don't read too much into the small speed differences between the Fourier and Hartley models.}

\textbf{Hybrid mixing-attention.} 
We choose two strong representative candidates from Table \ref{tab:pure_mixing}, namely the Hartley and Linear models, and replace a subset of the topmost mixing sublayers with self-attention. The results are summarized in Table \ref{tab:hybrid_mixing}.

\begin{table}
    \caption{Metrics for hybrid attention-mixing models. Hartley-$k$ denotes a model with $k$ self-attention sublayers and $12-k$ Hartley sublayers.}
    \label{tab:hybrid_mixing}
    \centering
    \setlength{\tabcolsep}{4pt}
    \begin{tabular}{l | c  c  c | c}
        \hline
          & \multicolumn{3}{c|}{Accuracy (\%)} & Speed \\ 
         Model & GLUE & MLM & NSP & (ms/batch) \\ \hline \hline
        %  BERT & 82.1 & \underline{63.8} & \textbf{81.8} & 300 \\ \hline
         Hartley-0 &   78.0 &	    58.5 &	74.9 & 	\textbf{172	(1.76x)} \\
         Hartley-1 &   78.0 &	    51.9 &	75.3 &	\underline{183	(1.66x)} \\ 
         Hartley-2 &   81.1 & 	    61.3 &	79.8 &	193	(1.57x) \\
         Hartley-3 &   77.9 & 	    50.3 &	76 &	204	(1.49x) \\ 
         Hartley-4 &   82.7 &   	62.6 &	81 &	216	(1.41x) \\
         Hartley-6 &   82.9 &    	63.5 &	\underline{81.2} &	234	(1.30x) \\ \hline
         Linear-0 &    77.7 &		57.6 &	77.4 &	200	(1.51x) \\
         Linear-1 &    78.1 &		62.5 &	78.3 &	208	(1.46x) \\
         Linear-2 &    82.8 &		62.8 &	81 &	218	(1.40x) \\
         Linear-3 &    82.8 &		63.3 &	81.6 &	226	(1.35x) \\
         Linear-4 $\star$ &    \underline{83.4} &		\underline{63.6} &	\textbf{81.7} &	235	(1.29x) \\
         Linear 6 &    \textbf{83.6} &		\textbf{64}   &	\textbf{81.7} &	251	(1.21x) \\ \hline
    \end{tabular} 
\end{table}

Once we include self-attention, we see that the hybrid Linear model offers larger quality gains than the hybrid Hartley model. Even though the hybrid Hartley models are faster, an iso-speed comparison still suggests that the hybrid Linear models are more efficient. For example, Hartley-6 and Linear-4 having roughly the same speed, but the Linear-4 model is more accurate. Hence, we opt to use the Linear-4 model. In Appendix \ref{subsec:extra_coord_descent} (Table \ref{tab:attention_layout}), we show that we get best accuracy when the self-attention sublayers at placed in the topmost layers.

\subsection{Model shape}
\label{subsec:shape_results}

All model shape experiments are run in parallel and start from the Linear-4 configuration.

\begin{table}
    \caption{Varying the model dimension, $d_{m}$. As in the Transformer, we set the model and embedding dimension to be equal. For the self-attention sublayers, we fix the number of self-attention heads to $d_{m}/64$.}
    \label{tab:model_dim}
    \centering
    \begin{tabular}{l | c  c  c | c}
        \hline
          & \multicolumn{3}{c|}{Accuracy (\%)} & Speed \\ 
         $d_{m}$ & GLUE & MLM & NSP & (ms/batch) \\ \hline \hline
         768 & \textbf{83.4}     &	 \textbf{63.6} & \textbf{81.7}	 & 235 (1.29x)	 \\
         512 $\star$ & \underline{83.0} & \underline{62.5}      &	\underline{80.9} 	 & 161 (1.89x)	 \\
         256 & 80.7     &	 58.9 &	 78.4 &	 \underline{91 (3.34x)} \\
         128 & 71.6      & 54	 &	73.8 &	\textbf{58 (5.29x)} \\ \hline
    \end{tabular} 
\end{table}

\textbf{Model dimensions.} In seeking a more efficient model, we attempt to slim our model down both by decreasing the model dimension (Table \ref{tab:model_dim}) and the intermediate MLP activation dimension (Table \ref{tab:feed_forward_dim} in Appendix \ref{subsec:extra_coord_descent}). For each coordinate, we find that there are cutoffs ($d_{ff}=2048$ and $d_{m}=512$) below which model quality drops drastically. We select these cutoffs as our optimal model shape values. It is in decreasing these two hyperparameters that we obtain the biggest speed-up in our model. However, there is a material degradation in quality that must be compensated by the increased capacity from the MoE sublayers in Section \ref{subsec:moe_results}. 

\textbf{Number of layers.} We vary the number of layers in Table \ref{tab:num_layers} in Appendix \ref{subsec:extra_coord_descent}. We opt for $14$ layers, beyond which we do not see quality gains. 
%  Because we plan to thin out our model, by decreasing $d_{ff}$ and $d_{m}$), we opt for slightly increasing the number of layers to 14.

\begin{table}
    \caption{Accuracy and speed metrics for Top-1 Tokens Choose (TC) and Experts Choose (EC) routing. 
    % We found that increasing $k>1$ (number of selected experts) for Tokens Choose routing yielded negligible quality gains over $k=1$.
    }
    \label{tab:routers}
    \centering
    \begin{tabular}{l | c  c  c | c}
        \hline
          & \multicolumn{3}{c|}{Accuracy (\%)} & Speed \\ 
         Router & GLUE & MLM & NSP & (ms/batch) \\ \hline \hline
         TC &    83.4   &		64 &	80.8 &	280 (1.09x) \\
         EC  $\star$ &    \textbf{83.5} &		\textbf{64.6} &	\textbf{81.2} &	283 (1.08x)\\ \hline
    \end{tabular} 
\end{table}

\subsection{MoE}
\label{subsec:moe_results}

Our starting configuration for our MoE ablations is the Linear-4 configuration with every other dense MLP sublayer replaced by an MoE sublayer ($6$ MoE sublayers) and $16$ experts in each MoE sublayer. We performed the MoE experiments in parallel to the model shape optimizations, so all MoE ablations are performed on a default Base sized model with $12$ layers, $d_{ff}=3072$ and $d_m=768$. 

As in \citep{zoph2022designing}, we find that we must adjust the fine-tuning learning protocol to better transfer any MoE MLM pre-training gains downstream. In particular, our MoE encoder models benefit from larger base learning rates ($\{10^{-4}, 5 \cdot 10^{-4}, 10^{-3}\}$) and larger dropout rates ($0.2$) for experts; see Appendix \ref{subsec:moe_finetuning} for a comparison of learning rates and expert dropout rates. For our final model comparison with BERT in Section \ref{sec:sparse_mixer}, we consider a wide range of base learning rates for all models.\footnote{Consistent with \citep{zoph2022designing}, we find that the larger base learning rates are not beneficial for the dense models.}

\textbf{Routers.} Routing mechanisms are compared in Table \ref{tab:routers}. We select Experts Choose routing as it obtains slightly higher accuracy results and does not require configuring a load balancing loss.

\begin{table}
    \caption{Varying the number and layout of MoE sublayers. Layout definition: 6-BOTTOM (first $6$ layers), 6-MIDDLE (middle $6$ layers) or 6-MIXED (every odd layer), 6-MIXED-odd (every even layer), and 6-TOP (final $6$ layers). 
    The number of experts and the expert capacity -- the number of tokens processed by each expert -- is fixed.
    Each MoE layer adds some compute and device communication overhead, slowing the model.}
    \label{tab:moe_layers}
    \centering
    \setlength{\tabcolsep}{2.5pt}
    \begin{tabular}{l | c  c  c | c}
        \hline
          & \multicolumn{3}{c|}{Accuracy (\%)} & Speed \\ 
         Config & GLUE & MLM & NSP & (ms/batch) \\ \hline \hline
        %  1-MIXED & 83.9     & 63.6	 & 81.4	 & 244	 \\
         2-MIXED & \underline{83.6}     & 63.6	 & 81.3	 & \textbf{246 (1.23x)}	 \\
         4-MIXED  $\star$ & \underline{83.6}     & 63.9 	 & 81.3	 & \underline{264 (1.15x)}	 \\
         6-MIXED & 83.5     & 64.6	 & 81.2	 & 283 (1.08x)	 \\
         12-MIXED & 83.1     & \underline{64.9}	 & 81.4	 & 352 (0.86x)	 \\ \hline
         6-BOTTOM  & 83.2 &	 62.7	 &	81.4 & 289 (1.05x) \\ 
         6-MIDDLE $\star$ & \textbf{83.9}     & 64	 &	\textbf{81.7} &	284 (1.08x) \\ 
         6-MIXED & 83.5    & 64.6	 &	81.2 & 283 (1.08x) \\ 
         6-MIXED-odd & 83.2      & 64.8	 & \underline{81.6}	 & 292 (1.04x)	 \\
         6-TOP & 83.4       & \textbf{65.4}	 & 81.2	 &	287 (1.06x)\\ \hline
    \end{tabular} 
\end{table}

\begin{table*}
    \caption{GLUE results on the \emph{Validation} split. We report F1/accuracy scores for QQP and MRPC, Spearman correlations for STS-B and accuracy scores for all other tasks. The MNLI accuracy metrics are reported by the match/mismatch splits.}
    \label{tab:glue}
    \centering
    \setlength{\tabcolsep}{5pt}
    \begin{tabular}{l| c c c c c c c c | c}
        \hline
         Model  & MNLI & QQP & QNLI & SST-2 & CoLA & STS-B & MRPC & RTE & Avg. \\ \hline \hline
         BERT & \textbf{81.3 / 81.8} & 86.7 / 90.3 & 88.9 & \textbf{91.1} & 77.6 & 87.3 & \textbf{90.5 / 86.8} & 69.7 & 84.7 \\
         Sparse Mixer & 80.7 / 81	&	\textbf{87.1 / 90.5}	&	\textbf{89.1}	&	90.9	&	\textbf{79}	&	\textbf{88.1}	&	90.4	 / 	86.3	&	\textbf{72.2}	&	\textbf{85.0} \\ \hline
    \end{tabular}
\end{table*}

\begin{table*}
    \caption{SuperGLUE \emph{Validation} results. We report macro-F1 scores for CB, micro-F1/exact match scores for MultiRC, F1/exact match scores for ReCoRD, and accuracy scores for all other tasks.}
    \label{tab:super_glue}
    \centering
    \begin{tabular}{l| c c c c c c c | c}
        \hline
         Model  & BoolQ & CB & COPA & MultiRC & ReCoRD & RTE & WiC & Avg. \\ \hline \hline
         BERT &  \textbf{74.6} & 86.4 / 85.7 & 58 & \textbf{74.1 / 26.2} & \textbf{68.6 / 52.2} & \textbf{65} & 65.7 & 65.7 \\
         Sparse Mixer & 74.4 & \textbf{93.3 / 92.9} & \textbf{62} & 72.4 / 22.5 & 65.9 / 49.2 & 64.6 & \textbf{66.5} & \textbf{66.4} \\ \hline
    \end{tabular}
\end{table*}

\textbf{MoE layers.} In Table \ref{tab:moe_layers}, we vary the number of MoE sublayers and the layout of those layers within the model. As we increase the number of MoE layers, MLM accuracy improves, but these pre-training gains do not always lead to better GLUE performance. This was a general trend that we observed; see also Appendix \ref{subsec:extra_coord_descent} and \citep{zoph2022designing}. 
% with abundant pre-training data, more MoE layers, and therefore larger model capacity, leads to stronger MLM results; however, on smaller fine-tuning datasets, there may be insufficient training signal to utilize the increased model capacity downstream.\footnote{This may be less of a concern for T5-style models \citep{raffel2019exploring} as they are simultaneously fine-tuned on all GLUE or SuperGLUE tasks.} 
We opt for 4 MoE layers, which performs well on GLUE and better than the 2 MoE sublayer model on the MLM task. 

The results of MoE layout experiments are clearer -- we opt to use the MIDDLE layout, placing all MoE sublayers in the middle layers of the model. Nevertheless, it is interesting to note that the TOP layout gives a big boost to MLM accuracy, but does not improve downstream GLUE accuracy. 
% It is possible that the TOP layout would perform better on a downstream task with a larger dataset.

\textbf{Number of experts.} We can increase the number of experts to increase the capacity of the model. For a large number of experts, the computational cost of the routing assignment is more significant, while the training signal to an individual expert becomes too weak to facilitate effective training as each expert processes too small a slice of data. Seeking a compromise between quality and speed, we ultimately opt to use 16 experts.
% \footnote{The optimal experts will vary based on hardware and the training dataset size.} 
Results are summarized in Table \ref{tab:experts} in Appendix \ref{subsec:extra_coord_descent}.

\textbf{Expert size.} We can control the number of parameters in each expert by varying its 
% intermediate activation dimension, 
$d_{ff}$. 
In Table \ref{tab:expert_size} (Appendix \ref{subsec:extra_coord_descent}), we find that: (1) using smaller experts yields a small accuracy drop, but limited speed benefits; and (2) increasing expert size increases MLM accuracy, 
% but the results do not always translate downstream to GLUE.
but not GLUE. So, for simplicity, we opt to keep the expert $d_{ff}$ the same size as the dense $d_{ff}$.
% , which we optimized in Section \ref{subsec:shape_results}.

\subsection{Sparse Mixer}
\label{sec:sparse_mixer_model}

Putting the preceding results 
% from our coordinate descent 
together, we arrive at the Sparse Mixer model in Figure \ref{fig:sparse_mixer_architecture}:
% \begin{align*}
% % \label{eq:sparse_mixer}
% \text{Shape}:&\ 14\ \text{layers},\ 512\ d_m,\ 2056\ d_{ff}; \nonumber \\
% \text{Sparse}:&\ 4\ \text{MIDDLE MoE},\ 16\ \text{experts},\\
% &\ 2056\ d_{ff},\ \text{EC routing},\ 1.0\ cf; \nonumber \\
% \text{Mixer}:&\ \text{Linear},\ 4\ \text{TOP Attention layers}. \nonumber
% \end{align*}
% \label{eq:sparse_mixer}
\\
\underline{Shape}: \ 14 \text{layers}, 512 $d_m$, 2056 $d_{ff}$; \\
\underline{Sparse}: 4 \text{MIDDLE MoE}, 16 {experts}, 2056 $d_{ff}$,
\text{\qquad\quad\ } {EC routing}, 1.0 $cf$; \\
\underline{Mixer}: \ \ {Linear}, 4 {TOP Attention layers}.

\begin{table}
    \caption{Model computational characteristics for BERT, Sparse Mixer (SM) and Fast Sparse Mixer (FSM), which is introduced in Section \ref{sec:sparse_mixer}. "Size" is the number of model parameters. Run speeds are measured by inference speed per example and pre-training step speed per example. 
    % Although they are "thinner" (small model dimension), Sparse Mixer models have more parameters than BERT because of their MoE layers.
    }
    \label{tab:vitals}
    \centering
    \setlength{\tabcolsep}{2pt}
    \begin{tabular}{l | c c | c  c}
        \hline
          & GFLOPS & Size & Inference & Training \\ 
         Model & (/ex) & (M) & (ms/ex) & (ms/ex) \\ \hline \hline
         BERT &  102 & 112 & 1.34 & 4.75 \\
         SM & {73} & 180 & {0.84 (1.61x)} & {2.87 (1.65x)}\\
         FSM & \textbf{60} & 180 & \textbf{0.68 (1.98x)} & \textbf{2.51 (1.89x)}\\ \hline
    \end{tabular} 
\end{table}
\begin{figure*}[tb]
    \centering
    \includegraphics[width=\textwidth]{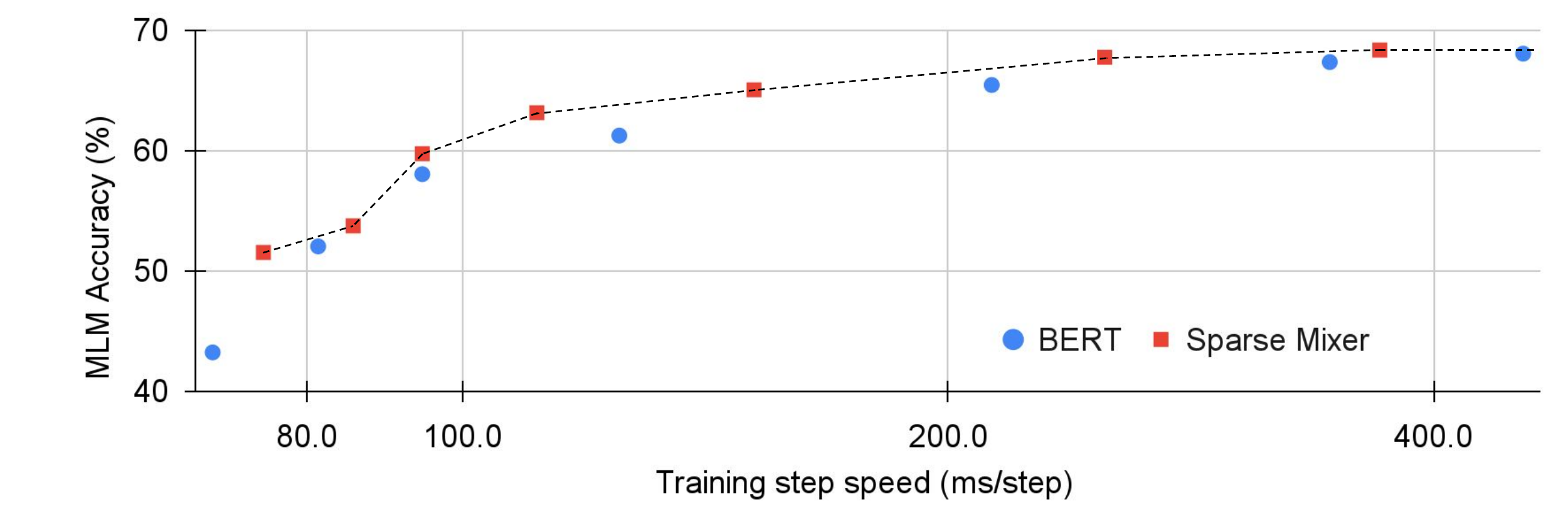}
    \caption{Pre-training Speed-accuracy trade-offs for Sparse Mixer and BERT. The corresponding model configurations are shown in Table \ref{tab:scale_sizes} in Appendix \ref{subsec:speed_nsp_accuracy}. The dashed line shows the Pareto efficiency frontier, indicating the best trade-offs. All models are trained on 32 TPU v3 chips. To better utilize the increased number of devices, we use a larger batch size of 256 but train for fewer (250k) steps.}
    \label{fig:mlm_scale}
\end{figure*}

\section{Evaluating Sparse Mixer}
\label{sec:sparse_mixer}

% \subsection{Full training comparison with BERT}
% \label{subsec:full_training}

\textbf{Full training comparison with BERT.} When comparing Sparse Mixer and BERT, both models are pre-trained on C4 for the full $1M$ steps, with batch size $64$, and then evaluated on both GLUE and SuperGLUE for a larger range of fine-tuning batch sizes (16, 32, and 64) and base learning rates ($\{10^{-5}, 5 \cdot 10^{-5}, 10^{-4}, 5 \cdot 10^{-4}, 10^{-3}\}$). The best results across all learning rates (for each task) and batch sizes (for all tasks) are shown in Tables \ref{tab:glue} and \ref{tab:super_glue}; see Appendix \ref{subsec:full_glue} for results for all batch sizes.\footnote{Following \citet{devlin2018bert}, we omit the WNLI task.} 
% Note that these results may be slightly different to the original BERT results \citep{devlin2018bert} as our models see fewer tokens during pre-training because of the smaller batch size ($64$ vs $256$). 

BERT and Sparse Mixer's GLUE scores are very similar, although they diverge a little more on SuperGLUE, where Sparse Mixer performs particularly well on the CB task, but underperforms BERT on the multi-label MultiRC and ReCoRD tasks. 
% Sparse Mixer performs better on COPA, although both models generally perform quite poorly.

% \subsection{Scaling the Sparse Mixer}
% \label{subsec:scaling}

\textbf{Scaling the Sparse Mixer.} Tables \ref{tab:glue}-\ref{tab:vitals} indicate that the Sparse Mixer is more efficient than BERT in the Base configuration. In Figure \ref{fig:mlm_scale}, we compare BERT and Sparse Mixer across a selection of model sizes. We use MLM accuracy as a proxy for model accuracy and pre-training step speed as a proxy for overall model speed. Pre-training step speed is a good proxy for inference speed (see Table \ref{tab:vitals}). MLM accuracy is only indicative of downstream accuracy. We construct an analogous speed-accuracy figure for NSP accuracy in Figure \ref{fig:nsp_scale} in Appendix \ref{subsec:speed_nsp_accuracy}.
These caveats aside, Figure \ref{fig:mlm_scale} suggests that Sparse Mixer's favorable speed and accuracy extends to other model sizes, as it defines the efficiency frontier across all model sizes considered.
% \footnote{We found that we had to increase the learning rate warm-up, from 2.5k steps to 5k steps, to stabilize BERT-Large. \citet{devlin2018bert} also note stability concerns for BERT-Large.}
 
% \subsection{Trading accuracy for more speed}
% \label{subsec:faster}

\begin{table}
    \caption{Fast Sparse Mixer (FSM). The default Sparse Mixer (SM) uses a capacity factor ($cf$) of 1 and a routing group size ($g$) of 4096. Several less favorable configurations are omitted.
    % : $cf=0.75$ was slower than $cf=0.5$ and yielded minimal quality gains; $g=1024$ further hurt accuracy; and
    % $cf=0.5$ with $g=2048$ 
    % yielded significant quality dips on SuperGLUE.
    }
    \label{tab:faster_sparse_mixers}
    \centering
    \setlength{\tabcolsep}{1pt}
    \begin{tabular}{l | c  c  | c}
        \hline
          & \multicolumn{2}{c|}{Accuracy (\%)} & Speed \\ 
         Model & GLUE & SuperGLUE & (ms/batch) \\ \hline \hline
         BERT & \underline{84.7}     &	\underline{65.7}  & 304	 \\
         SM & \textbf{85.0}  & \textbf{66.4} & 184 (1.65x)	 \\
         FSM ($cf$=$0.5$) &  \underline{84.7} & 65.6 &\textbf{161 (1.89x)} \\
         $g$=$2048$ & 84.5  & 65.1 & 173 (1.75x) \\
        %  $g$=$1024$ & 84.1 & 64.0 & 167 (1.82x) \\
         $cf$=$0.75$,$g$=$2048$  & 84.3  & 65.2  & \underline{165 (1.84x)} \\
         \hline
    \end{tabular} 
\end{table}

\begin{table}
    \caption{Stability of BERT, sparse BERTs and Sparse Mixer (SM). BERT-$k$ denotes a BERT model with $k$ MoE layers.
    The "unstable" runs experience gradient blow-up and fail to converge to an optimal loss (or converge at all).
    We use batch sizes of 64 and 256.
    % ; for reference, \citet{devlin2018bert} uses 256. 
    Accuracy and speed metrics are reported for 64 batch runs.}
    \label{tab:stability}
    \centering
    \setlength{\tabcolsep}{2.pt}
    \begin{tabular}{l | c c | c  c | c}
        \hline
         & \multicolumn{2}{c|}{Stable} & \multicolumn{2}{c|}{Accuracy (\%)} & Speed \\ 
         Model & 64 & 256 & GLUE & S.GLUE & (ms/batch) \\  \hline \hline
         BERT & 3/4 & \textbf{4/4} & 84.7& 65.7 & 304	 \\
         SM & \textbf{4/4} & \textbf{4/4} & \textbf{85.0}  & \textbf{66.4} & \textbf{184 (1.65x)}	 \\
         BERT-4 & 0/4 & 0/4 & - & - & - \\
         BERT-12 & 1/4 & 0/4 & 84.1 & 60.9 & 426 (0.71x) \\ \hline
    \end{tabular} 
\end{table}

\textbf{Trading accuracy for more speed.} We design an even sparser model by decreasing the expert capacity factor. This decreases the number of tokens that each expert processes and yields significant speed-ups for a limited quality degradation: for a minor (0.2\%) accuracy drop on SuperGLUE relative to BERT, a Sparse Mixer with capacity factor of 0.5 trains 89\% faster and runs inference 98\% faster; see Table \ref{tab:vitals}. We name this variant of the model Fast Sparse Mixer.\footnote{The minimal accuracy drop from decreasing the capacity factor, $cf$, could potentially be mitigated by training with the default $cf$=$1$, and then using the smaller $cf$ during inference, although the mismatch may yield unexpected results.} We also experiment with decreasing the token routing group size, but this leads to larger quality drops.

% \subsection{Stability}
% \label{subsec:sparse_bert}

\textbf{Stability.} Table \ref{tab:stability} compares the stability of Sparse Mixer, BERT and "sparse BERTs" -- MoE variants of BERT.  Sparse Mixer is very stable, even relative to (dense) BERT. The sparse BERTs are highly unstable, with only one stable run that ultimately yields a slow model that significantly underperforms BERT.\footnote{It may be possible to improve the stability of the sparse BERTs by adjusting the MoE configurations, such as the number of experts or the router z-loss. That said, as detailed in Section \ref{subsec:moe_results}, Sparse Mixer was robust under all such changes.}
%  It is possible that with more trials, we will eventually find a BERT-4 model that is also trains successfully, but working with a model which trains robustly only $\leq25\%$ of the time is not practical.
%
% Stability aside, even if the sparse BERT models could be made more stable, they would inevitably be slower than BERT.
We hypothesize that the Sparse Mixer's improved stability is due to replacing most of the self-attention sublayers with mixing, which constrains the model to a less variable mixing basis. 

\section{Conclusions}
\label{sec:conclusion}

Mixing transformations and MoE play well together. Utilizing MoE for capacity and mixing for speed and stability, we introduced the Sparse Mixer -- a model that slightly 
% ($<1\%$) 
\emph{outperforms} BERT on GLUE and SuperGLUE, but more importantly trains 65\% faster and runs inference 61\% faster. We also presented a faster variant, Fast Sparse Mixer, that marginally 
% ($<0.2\%$) 
\emph{under-performs} BERT on SuperGLUE, but trains and runs nearly twice as fast: 89\% faster training and 98\% faster inference. 
% We justified the design of these two models by carefully ablating through various mixing mechanisms, MoE configurations, and model hyperparameters.
Sparse Mixer overcomes many of the speed and stability concerns of MoE models and offers the prospect of serving sparse student models.

% We have restricted our focus to encoders. In future work, we plan to investigate porting our recipes to T5 models \citep{raffel2019exploring}. Sparse mixer encoder-decoder and decoder-only models are, in principle, straightforward extensions: Linear decoders can be designed by ``causally'' masking the Linear matrix and encoder-decoder mixing can also be designed with careful masking. However, we suspect that parts of the coordinate descent program will need to be repeated. For example, evidence suggests that cross-attention may be crucial to performance \citep{you2020hard} of Transformer-like encoder-decoder models. 
% % We do remark that the sparse mixer encoder could be used as a drop in replacement in a Transformer as other works have successfully demonstrated; see, for example, \cite{zaheer2020big, guo2021longt5}.

% There are also a number of orthogonal future directions that could be adapted to potentially improve the training regime of the Sparse Mixer, such as training for much longer as for RoBERTa \citep{liu2019roberta} or using the ELECTRA generator-discriminator training setup \citep{clark2020electra}.

\section*{Limitations}
\label{sec:limitations}

\textbf{Encoder only model.} 
% TODO(JLT)
We have focused our work on BERT-like models as they are extremely wide used.\footnote{See, for example, \url{https://huggingface.co/models}.} 
However, this limits our focus to encoders, which are not suitable for generative tasks. 
% We plan to investigate porting our recipes to T5 models \citep{raffel2019exploring}. 
Sparse mixer encoder-decoder and decoder-only models are, in principle, straightforward extensions: Linear decoders can be designed by ``causally'' masking the Linear matrix and encoder-decoder mixing can also be designed with careful masking. However, we suspect that parts of the coordinate descent program will need to be repeated. For example, evidence suggests that cross-attention may be crucial to performance of encoder-decoder models \citep{you2020hard}. 
Nevertheless, we hope that the current Sparse Mixer recipe acts as a starting point and a roadmap for generalizing to other architectures.

\textbf{More diverse tasks and learning frameworks.} We only evaluated Sparse Mixer on GLUE and SuperGLUE. It would be good to look at broader set of tasks, including Q\&A. We also stuck to the original BERT training setup \citep{devlin2018bert}, but there are potential training regime improvements that could be introduced,  such as training for much longer as for RoBERTa \citep{liu2019roberta} or using the ELECTRA generator-discriminator training setup \citep{clark2020electra}.

\textbf{"Manual ML".} In designing the Sparse Mixer architecture, we have optimized the model configuration one hyperparameter coordinate at a time. So while our manual gradient descent offers interpretability and pedagogical insight, it is potentially sub-optimal. It would be exciting to see future work expand both the coordinate space and jointly optimize multiple coordinates using Automated Machine Learning (AutoML) \citep{thornton2013auto, liu2018darts, peng2020pyglove}.

% \textbf{Limited model sizes.} We didn't scale Sparse Mixer beyond Large -- the largest size considered in \citep{devlin2018bert}. It would be good to see how Sparse Mixer scales to sizes comparable to T5 XL or XXL \citep{raffel2019exploring}.

\textbf{Long input sequences.} Because of the presence of attention layers, Sparse Mixer will not scale as well as efficient Transformers to long sequence inputs. This could be compensated by dropping in efficient approximations of the attention mechanism \cite{tay2020long}.

\section*{Acknowledgements}
We would like to give a big thanks to Parker Schuh for critical help with the Mixture-of-Experts implementation. We also thank Santiago Onta\~{n}\'{o}n for many helpful brainstorming sessions around the design and evaluation of the Sparse Mixer model.

\bibliographystyle{acl_natbib}
\bibliography{anthology, references}

\appendix
\clearpage

\section{Appendices}
\label{sec:appendix}

\subsection{Base architecture}
\label{subsec:bert_arch}

\begin{figure}
    \centering
    \includegraphics[width=0.45\textwidth]{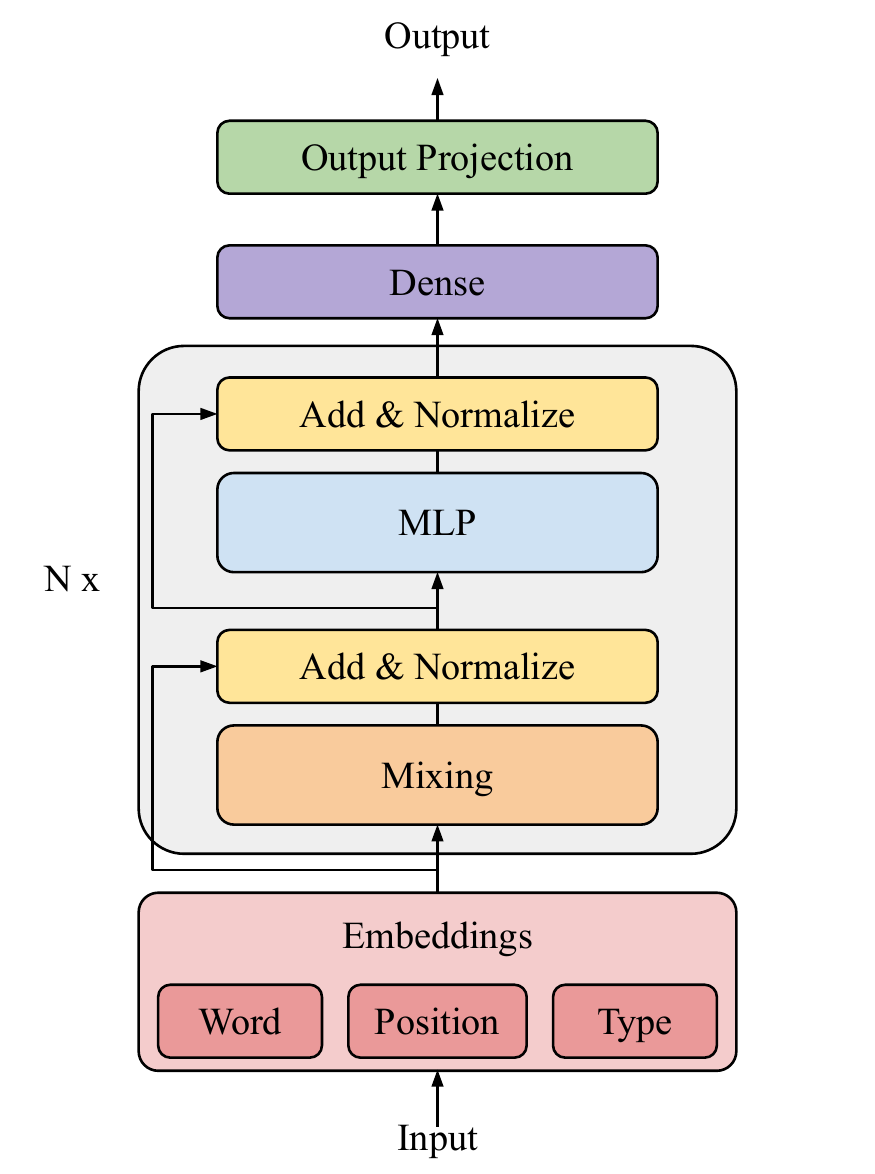}
    \caption{Block based encoder architecture. The model has $N$ encoder blocks, each containing mixing and MLP sublayers. Each MLP sublayer may be sparse or dense. Each mixing sublayer may use self-attention or a mixing transformation.}
    \label{fig:general_architecture}
\end{figure}

Our design space for the Sparse Mixer builds off of the stacked encoder blocks of BERT \citep{devlin2018bert} in Figure \ref{fig:general_architecture}. Each encoder block contains a mixing sublayer and an MLP sublayer, connected with residual connections and layer norms. We keep the standard BERT input embedding and output projection layers \citep{devlin2018bert}.

\begin{figure*}
    \centering
    \includegraphics[width=\textwidth]{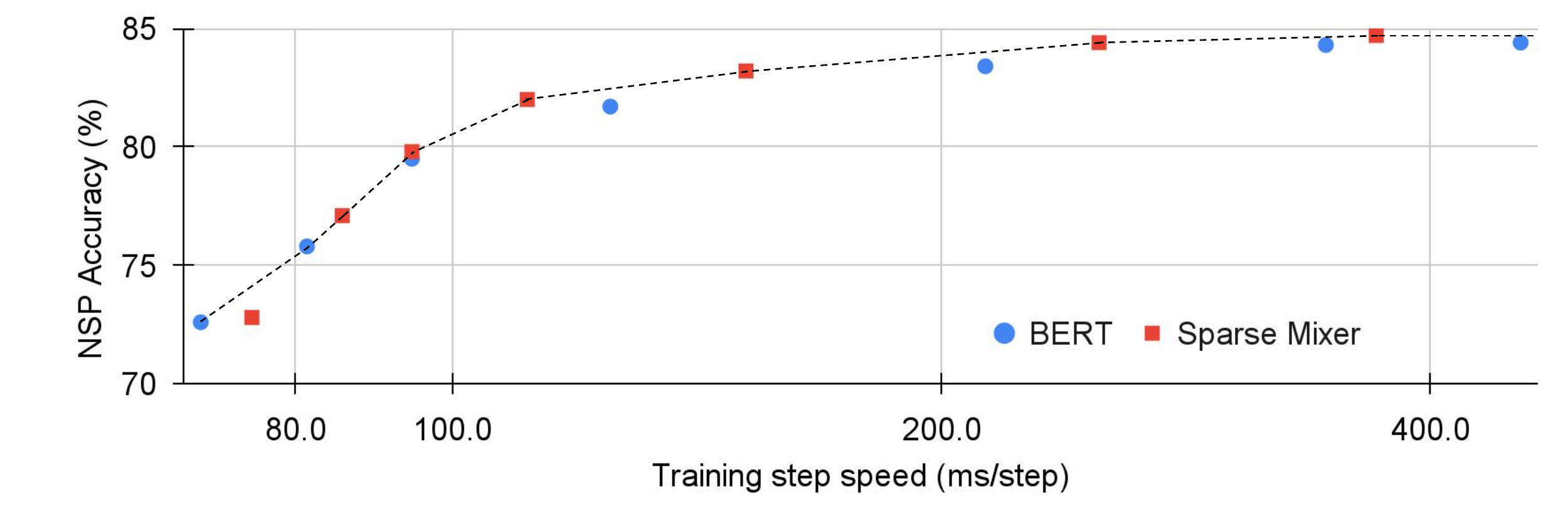}
    \caption{NSP pre-training Speed-accuracy trade-offs for Sparse Mixer and BERT. The dashed line shows the Pareto efficiency frontier, indicating the best trade-offs.}
    \label{fig:nsp_scale}
\end{figure*}

\subsection{Exploring more coordinates}
\label{subsec:extra_coord_descent}

\begin{table}
    \caption{Average accuracy metrics and median pre-training step speeds for varying where the self-attention sublayer are placed within the Linear-4 model. Speed-ups relative to BERT (see Table \ref{tab:vitals}) are shown in parentheses. The best metrics are highlighted in boldface. The star indicate the selected configuration.}
    \label{tab:attention_layout}
    \centering
    \setlength{\tabcolsep}{4pt}
    \begin{tabular}{l | c  c  c | c}
        \hline
          & \multicolumn{3}{c|}{Accuracy (\%)} & Speed \\ 
         Layout & GLUE & MLM & NSP & (ms/batch) \\ \hline \hline
         BOTTOM &    82   &		62.6 &	81 &	236	(1.29x) \\
         MIDDLE &    82.4 &		63.2 &	80.7 &	236	(1.29x) \\
         MIXED &    82.7  &		\textbf{63.6} &	81.1 &	235	(1.29x) \\
         TOP $\star$ &    \textbf{83.4} &		\textbf{63.6} &	\textbf{81.7} &	235	(1.29x) \\ \hline
    \end{tabular} 
\end{table}

\textbf{Attention sublayer layout.} Where should the $4$ attention sublayers be placed within the model? We check whether it best to place the $4$ self-attention sublayers at the TOP (final $4$ layers), BOTTOM (first $4$ layers), MIDDLE (middle $4$ layers) or MIXED (every third layer). Table \ref{tab:attention_layout} shows that the TOP layout is best. 

\textbf{Mixing dead ends.} We tested two other mixing modifications that yielded no quality (or latency) gains: (1) adding a bias term to the mixing transformations, and (2) adding dropout during fine-tuning to the mixing sublayers.

\textbf{Intermediate activation dimension.} Below $d_{ff}=2048$, the model quality drops significantly. We select this cutoff as our optimal model MLP dimension.

\textbf{Number of layers.} We do not see quality gains beyond $14$ layers. Because we plan to thin out our model (decrease $d_{ff}$ and $d_{m}$), we opt for slightly increasing the number of layers to 14.

\begin{table}
    \caption{Metrics for various intermediate MLP activation dimensions, $d_{ff}$.}
    \label{tab:feed_forward_dim}
    \centering
    \begin{tabular}{l | c  c  c | c}
        \hline
          & \multicolumn{3}{c|}{Accuracy (\%)} & Speed \\ 
         $d_{ff}$ & GLUE & MLM & NSP & (ms/batch) \\ \hline \hline
         3072 & \underline{83.4}     &	 \textbf{63.6} & \textbf{81.7}	 & 235 (1.29x)	 \\
         2560 & 83.2      &	\underline{62.2}  & \underline{81.1}	 & 208 (1.46x) \\
         2048 $\star$ & \textbf{83.5}      &	62  & 81	 & 189 (1.61x)	 \\
         1024 & 82.9     &	60.7 &	 80.6 &	 \underline{146 (2.09x)} \\
         768 & 82.8      & 60.4	 &	80.7 &	\textbf{135 (2.24x)} \\ \hline
    \end{tabular} 
\end{table}

\begin{table}
    \caption{Varying the number of model layers. The results are for post-layer normalization, as in BERT. We obtained similar results for pre-layer normalization.}
    \label{tab:num_layers}
    \centering
    \begin{tabular}{l | c  c  c | c}
        \hline
          & \multicolumn{3}{c|}{Accuracy (\%)} & Speed \\ 
         Layers & GLUE & MLM & NSP & (ms/batch) \\ \hline \hline
         6 & 82.3     &	 62.4 & 80.9	 & \textbf{142 (2.14x)}	 \\
         10 & \underline{83.6}      &	63.4  & 81.4	 & \underline{201 (1.51x)}	 \\
         12 & 83.4     &	 63.6 &	 \textbf{81.7} &	 235 (1.29x) \\
         14 $\star$ & \textbf{83.8}      & \textbf{64}	 &	\underline{81.6} & 260 (1.17x) \\ 
         18 & \textbf{83.8}      & \underline{63.9}	 &	81.5 &	320 (0.95x) \\ \hline
    \end{tabular} 
\end{table}

\textbf{Number of experts.} In Table \ref{tab:experts}, we increase the number of experts to increase the capacity of the model. As discussed in Section \ref{subsec:moe_background}, we simultaneously decrease each expert's capacity (the number of tokens it processes) to prevent FLOPS from growing. We suspect that, for a large number of experts, the training signal to an individual expert becomes too weak to facilitate effective training as each expert processes too small a slice of data. This is particularly apparent on downstream tasks where there are fewer training examples. Furthermore, for a large number of experts ($\geq64$), the computational cost of the routing assignment starts to become significant. Seeking a compromise between quality and speed, we ultimately opt to use 16 experts.\footnote{The optimal experts will vary based on hardware and the training dataset size.}

\begin{table}
    \caption{Increasing the number of experts. These experiments were run in parallel to those of Table \ref{tab:moe_layers} and therefore use the default 6-MIXED setup. Speed-ups relative to BERT (see Table \ref{tab:vitals}) are shown in parentheses. Based on the speed slowdown for 64 experts and the relatively weak performance of 32 experts, we did not evaluate 64 experts on GLUE.}
    \label{tab:experts}
    \centering
    \setlength{\tabcolsep}{5pt}
    \begin{tabular}{l | c  c  c | c}
        \hline
          & \multicolumn{3}{c|}{Accuracy (\%)} & Speed \\ 
         Experts & GLUE & MLM & NSP & (ms/batch) \\ \hline \hline
         8 & 83.3      & 64.1	 & 81.2	 & \textbf{277 (1.10x)}	 \\
         16 $\star$ & \textbf{83.5}     & 64.6	 & \underline{81.2}	 & \underline{283 (1.08x)}	 \\
         32 & 83.3     & \underline{65.1}	 & \textbf{81.3}	 & 292 (1.04x)	 \\
         64 & -     & \textbf{65.4}	 & 80.9	 & 327 (0.93x)	 \\ \hline
    \end{tabular} 
\end{table}

\textbf{Expert size.} The number of parameters in each expert can be controlled by varying its intermediate activation dimension, $d_{ff}$. If we can maintain accuracy while decreasing the size of each expert, that will speed up our model. On the other hand, if we can achieve large accuracy gains by increasing the size of experts, we can potentially use that to offset shrinking the rest of the model; for example, by constructing a "thin", fast model with only a few "heavy", high capacity MoE layers. In Table \ref{tab:expert_size}, we see that neither of these scenarios plays out cleanly: (1) Using smaller experts ($d_{ff}=1536$) yields only a small accuracy drop, but little speed benefit. (2) As was the case for increasing the number of MoE sublayers and the number of experts, increasing expert size increases MLM accuracy, but the results do not translate downstream to GLUE. For simplicity, we opt to keep the expert $d_{ff}$ the same size as the dense $d_{ff}$, which is optimized in Section \ref{subsec:shape_results}.

\begin{table}
    \caption{Varying the size of experts by varying each expert's intermediate activation dimension, $d_{ff}$. These experiments were run in parallel to those of Table \ref{tab:moe_layers} and therefore use the default 6-MIXED setup.}
    \label{tab:expert_size}
    \centering
    \begin{tabular}{l | c  c  c | c}
        \hline
          & \multicolumn{3}{c|}{Accuracy (\%)} & Speed \\ 
         $d_{ff}$ & GLUE & MLM & NSP & (ms/batch) \\ \hline \hline
         1536 & \underline{83.4}     &	 64 & 81.1	 & \textbf{274 (1.11x)}	 \\
         3072 $\star$ & \textbf{83.5}      &	64.6  & 81.2	 & \underline{283 (1.08x)}	 \\
         6144 & 83.3     &	 \underline{65.3} &	 \underline{81.5} &	 305 (1.00x)\\
         12288 & 83.1      & \textbf{65.9}	 &	\textbf{81.7} &	350 (0.87x) \\ \hline
    \end{tabular} 
\end{table}

\textbf{MoE dead ends.} (1) Changing the expert nonlinearity from GELU to RELU had little effect on downstream performance. Changing the nonlinearity to GEGLU (GELU Gated Linear Units) \citep{shazeer2020glu} only slowed down the model.\footnote{A limitation of our GEGLU investigation is that we did not simultaneously shrink the intermediate activation dimension to account for the extra activation function, as in \citep{shazeer2020glu, narang2021transformer}.} (2) Although the router z-loss had little effect on model performance, we included it for potential stability benefits. (3) \citet{fedus2021switch} recommend using a smaller scaled weight initialization to provide stability to MoE encoder-decoder models, especially in larger configurations. However, we obtained the best results, for our encoder-only model, with BERT's default kernel initialization.

\begin{table*}
    \caption{Full GLUE results (\emph{Validation} split) for all coordinate descent experiments. See the corresponding table for descriptions of each configuration. For dense models (Tables \ref{tab:pure_mixing}, \ref{tab:hybrid_mixing}, \ref{tab:attention_layout}, \ref{tab:feed_forward_dim}, \ref{tab:model_dim} and \ref{tab:num_layers}), we report the best scores across $\{10^{-5},5\cdot10^{-5},10^{-4}\}$ base learning rates, while for MoE models (Tables \ref{tab:routers}, \ref{tab:moe_layers}, \ref{tab:experts} and \ref{tab:expert_size}), we use $\{10^{-4},5\cdot10^{-4},10^{-3}\}$. We report F1/accuracy scores for QQP and MRPC, Spearman correlations for STS-B and accuracy scores for all other tasks. The MNLI accuracy metrics are reported by the match/mismatch splits. The top two average scores for each experiment set are boldfaced/underlined.}
    \label{tab:glue_coord}
    \small
    \centering
    \begin{tabular}{c | l | c c c c c c c c | c}
        \hline
         Table & Model  & MNLI & QQP & QNLI & SST-2 & CoLA & STS-B & MRPC & RTE & Avg. \\ \hline \hline
         
    &	Fourier	&	74.5	/	75.6	&	84.3	/	88.3	&	82.2	&	88.5	&	69.6	&	80.2	&	83.5	/	73.8	&	62.1	&	\textbf{78.4}	\\	
	&	Hartley	&	75	/	75.7	&	84.4	/	88.1	&	82.6	&	87.7	&	69.4	&	80.9	&	82.2	/	72.1	&	59.9	&	\underline{78.0}	\\	
\ref{tab:pure_mixing}	&	Circulant	&	69.7	/	70.9	&	83.3	/	87.6	&	76	&	89.2	&	76.1	&	56.1	&	82.7	/	71.8	&	62.8	&	75.1	\\	
	&	Toeplitz	&	73.2	/	73.5	&	84.2	/	88.3	&	78.1	&	88.6	&	73	&	66.1	&	82.6	/	71.6	&	62.8	&	76.5	\\	
	&	Linear	&	73.4	/	74.3	&	84.3	/	88.2	&	80.2	&	89.6	&	74.6	&	68.2	&	83.5	/	75.2	&	63.5	&	77.7	\\	\hline
																												
	&	Hartley-0	&	75	/	75.7	&	84.4	/	88.1	&	82.6	&	87.7	&	69.4	&	80.9	&	82.2	/	72.1	&	59.9	&	78.0	\\	
	&	Hartley-1	&	72.7	/	73.6	&	83.9	/	87.8	&	82.1	&	86.8	&	71.1	&	79.5	&	82.8	/	75.7	&	62.1	&	78.0	\\	
	&	Hartley-2	&	79.5	/	78.3	&	86	/	89.7	&	86.2	&	89.3	&	74	&	85.7	&	84.6	/	76.7	&	62.1	&	81.1	\\	
	&	Hartley-3	&	74.2	/	75	&	83.5	/	87.9	&	81.9	&	88.8	&	70.4	&	78.3	&	83.1	/	73.3	&	61	&	77.9	\\	
	&	Hartley-4	&	79	/	80.2	&	86.7	/	90.1	&	87.2	&	89.9	&	75.2	&	86.1	&	87.5	/	82.4	&	65	&	82.7	\\	
\ref{tab:hybrid_mixing}	&	Hartley-6	&	80	/	81	&	86.7	/	90.2	&	88	&	90.7	&	75.3	&	86.5	&	87.3	/	81.9	&	64.6	&	82.9	\\	
	&	Linear-0	&	73.4	/	74.3	&	84.3	/	88.2	&	80.2	&	89.6	&	74.6	&	68.2	&	83.5	/	75.2	&	63.5	&	77.7	\\	
	&	Linear-1	&	74.4	/	74.9	&	84.9	/	88.8	&	81.8	&	91.4	&	78	&	69	&	83.4	/	73	&	59.9	&	78.1	\\	
	&	Linear-2	&	80.1	/	80.5	&	87	/	90.3	&	87.6	&	90	&	78.2	&	86.7	&	85.3	/	78.2	&	66.8	&	82.8	\\	
	&	Linear-3	&	80.2	/	81	&	87	/	90.4	&	87.3	&	90.5	&	77.9	&	87.8	&	84.7	/	78.4	&	66.1	&	82.8	\\	
	&	Linear-4	&	80.4	/	81.2	&	87.2	/	90.4	&	88	&	91.3	&	76.7	&	87.4	&	87.2	/	81.6	&	65.7	&	\underline{83.4}	\\	
	&	Linear-6	&	80.7	/	81.6	&	87.3	/	90.6	&	87.8	&	90.5	&	78.2	&	87.2	&	88.7	/	83.6	&	63.9	&	\textbf{83.6}	\\	\hline
																												
	&	BOTTOM	&	79.1	/	79.6	&	86.8	/	90.3	&	86.2	&	90.7	&	73.2	&	85.5	&	87.5	/	82.1	&	61	&	82.0	\\	
\ref{tab:attention_layout}	&	MIDDLE	&	80	/	80.3	&	86.7	/	90.3	&	86.5	&	89.2	&	72.2	&	86.3	&	88.3	/	83.1	&	63.5	&	82.4	\\	
	&	MIXED	&	79.5	/	80.3	&	87.1	/	90.5	&	87.6	&	90.7	&	75	&	85	&	87.1	/	82.1	&	64.6	&	82.7	\\	
	&	TOP	&	80.4	/	81.2	&	87.2	/	90.4	&	88	&	91.3	&	76.7	&	87.4	&	87.2	/	81.6	&	65.7	&	83.4	\\	\hline
																												
	&	$d_{ff}$=3072	&	80.4	/	81.2	&	87.2	/	90.4	&	88	&	91.3	&	76.7	&	87.4	&	87.2	/	81.6	&	65.7	&	\underline{83.4}	\\	
	&	$d_{ff}$=2560	&	80.3	/	81.5	&	87.2	/	90.5	&	88.3	&	91.6	&	77.8	&	86.9	&	85.5	/	79.2	&	66.1	&	83.2	\\	
\ref{tab:feed_forward_dim}	&	$d_{ff}$=2048	&	80.4	/	81.1	&	87.1	/	90.5	&	87.7	&	91.3	&	76.6	&	87	&	87.8	/	82.4	&	66.8	&	\textbf{83.5}	\\	
	&	$d_{ff}$=1024	&	79.9	/	80.8	&	87	/	90.3	&	87.3	&	90	&	76.3	&	87.4	&	87.7	/	81.9	&	62.8	&	82.9	\\	
	&	$d_{ff}$=768	&	80.1	/	81.1	&	86.7	/	90.2	&	87.4	&	89.7	&	75.7	&	86.6	&	87.7	/	82.6	&	62.8	&	82.8	\\	\hline
																												
	&	$d_{m}$=768	&	80.4	/	81.2	&	87.2	/	90.4	&	88	&	91.3	&	76.7	&	87.4	&	87.2	/	81.6	&	65.7	&	\textbf{83.4}	\\	
\ref{tab:model_dim}	&	$d_{m}$=512	&	80.1	/	80.9	&	86.6	/	90.1	&	87.4	&	90.8	&	77	&	87.2	&	86.3	/	80.1	&	66.4	&	\underline{83.0}	\\	
	&	$d_{m}$=256	&	77.9	/	78.5	&	84.4	/	88.1	&	85.2	&	89.8	&	73.9	&	83.8	&	84.5	/	76	&	66.1	&	80.7	\\	
	&	$d_{m}$=128	&	74.9	/	75.9	&	84.2	/	88.1	&	84	&	88	&	69.8	&	9.7	&	82.1	/	70.3	&	60.3	&	71.6	\\	\hline
																												
	&	6 layers	&	79.9	/	81	&	87	/	90.4	&	87.3	&	90	&	76	&	86.7	&	85.1	/	76.5	&	65	&	82.3	\\	
	&	10 layers	&	80.7	/	81.6	&	87.1	/	90.4	&	87.5	&	87.5	&	76.5	&	87.2	&	90.2	/	86.3	&	64.3	&	83.6	\\	
\ref{tab:num_layers} 	&	12 layers	&	80.4	/	81.2	&	87.2	/	90.4	&	88	&	91.3	&	76.7	&	87.4	&	87.2	/	81.6	&	65.7	&	83.4	\\	
	&	14 layers	&	80.5	/	81.7	&	87.2	/	90.6	&	88.1	&	90.4	&	78.4	&	87.8	&	87.2	/	81.4	&	68.2	&	\textbf{83.8}	\\	
	&	18 layers	&	81.1	/	81.9	&	87.1	/	90.5	&	87.7	&	90.6	&	79.2	&	87.3	&	87	/	82.6	&	67.2	&	\textbf{83.8}	\\	\hline
																												
\ref{tab:routers}	&	Tokens Choose	&	80.2	/	81.2	&	86.7	/	90.2	&	87.8	&	90.1	&	77.4	&	87.3	&	87.6	/	82.4	&	66.4	&	83.4	\\	
	&	Experts Choose	&	80.4	/	81.1	&	86.8	/	90.2	&	87.5	&	90.6	&	75.9	&	84.9	&	88.7	/	83.6	&	68.6	&	\textbf{83.5}	\\	\hline
																												
	&	2-MIXED	&	80.6	/	81.3	&	87.1	/	90.5	&	87.1	&	90.9	&	76.7	&	86.6	&	88	/	82.4	&	68.2	&	\underline{83.6}	\\	
	&	4-MIXED	&	80.7	/	80.8	&	86.7	/	90.2	&	87.3	&	90.8	&	76.8	&	87.5	&	88	/	82.4	&	68.2	&	\underline{83.6}	\\	
	&	6-MIXED	&	80.4	/	81.1	&	86.8	/	90.2	&	87.5	&	90.6	&	75.9	&	84.9	&	88.7	/	83.6	&	68.6	&	83.5	\\	
	&	12-MIXED	&	80	/	80.2	&	86.7	/	90.1	&	86.3	&	90.7	&	75.5	&	87.5	&	87.7	/	82.1	&	66.8	&	83.1	\\	
\ref{tab:moe_layers}	&	6-BOTTOM	&	79.9	/	81	&	86.9	/	90.3	&	86.9	&	90.3	&	76.1	&	87.1	&	87.9	/	82.6	&	66.4	&	83.2	\\	
	&	6-MIDDLE	&	80.3	/	80.9	&	90.6	/	87.2	&	87.6	&	90.7	&	77.6	&	86.9	&	89.7	/	85.3	&	66.4	&	\textbf{83.9}	\\	
	&	6-MIXED	&	80.4	/	81.1	&	86.8	/	90.2	&	87.5	&	90.6	&	75.9	&	84.9	&	88.7	/	83.6	&	68.6	&	83.5	\\	
	&	6-MIXED$\ast$	&	80.2	/	81.4	&	87	/	90.4	&	87.6	&	90.8	&	75.5	&	86.6	&	88.2	/	82.8	&	65.0	&	83.2	\\	
	&	6-TOP	&	80.5	/	81.1	&	86.5	/	90	&	87.6	&	90.5	&	77.3	&	85.8	&	87.7	/	82.8	&	67.1	&	83.4	\\	\hline
																												
	&	8 experts	&	80.4	/	81.4	&	86.2	/	89.8	&	88	&	90.9	&	75.7	&	86.7	&	88.3	/	83.1	&	65.7	&	83.3	\\	
\ref{tab:experts}	&	16 experts	&	80.4	/	81.1	&	86.8	/	90.2	&	87.5	&	90.6	&	75.9	&	84.9	&	88.7	/	83.6	&	68.6	&	\textbf{83.5}	\\	
	&	32 experts	&	80.4	/	81.4	&	86.8	/	90.2	&	87.4	&	91.1	&	78.3	&	86.7	&	87	/	81.4	&	65.3	&	83.3	\\	\hline
																												
	&	$d_{ff}$=1536	&	80.6	/	81.3	&	86.8	/	90.2	&	87.2	&	90.6	&	76.5	&	85.5	&	87.7	/	82.4	&	68.6	&	\underline{83.4}	\\	
\ref{tab:expert_size}	&	$d_{ff}$=3072	&	80.4	/	81.1	&	86.8	/	90.2	&	87.5	&	90.6	&	75.9	&	84.9	&	88.7	/	83.6	&	68.6	&	\textbf{83.5}	\\	
	&	$d_{ff}$=6144	&	80.3	/	81.4	&	87.1	/	90.4	&	87.8	&	90.6	&	76.4	&	87.3	&	87.3	/	80.9	&	67.1	&	83.3	\\	
	&	$d_{ff}$=12288	&	80.7	/	81.9	&	85.9	/	89.7	&	87.7	&	90.9	&	77.9	&	87.2	&	85.6	/	79.7	&	67.1	&	83.1	\\	\hline
    \end{tabular}
\end{table*}

\begin{table*}
    \caption{GLUE results (\emph{Validation} split) for final comparison of BERT, Sparse Mixer (SM), Fast Sparse Mixer (FSM) and other variants for different batch sizes. See the corresponding table for descriptions of each configuration. For each task, we select the best result across the base learning rates $\{10^{-5}, 5 \cdot 10^{-5}, 10^{-4}, 5 \cdot 10^{-4}, 10^{-3}\}$. The highest average score for each model (across the three batch sizes) is highlighted in boldface.}
    \label{tab:glue_batch}
    \small
    \setlength{\tabcolsep}{5pt}
    \centering
    \begin{tabular}{c | c | c | c c c c c c c c | c}
        \hline
         Table & Model & Batch  & MNLI & QQP & QNLI & SST-2 & CoLA & STS-B & MRPC & RTE & Avg. \\ \hline \hline
         
    &		&	16	&	81.3	/	81.8	&	86.7	/	90.3	&	88.9	&	91.1	&	77.6	&	87.3	&	90.5	/	86.8	&	69.7	&	\textbf{84.7}	\\	
	&	BERT	&	32	&	81.2	/	81.5	&	86.9	/	90.3	&	88.6	&	91.3	&	77.6	&	87.4	&	90.4	/	86.5	&	70	&	\textbf{84.7}	\\	
\ref{tab:super_glue}	&		&	64	&	81.3	/	81.4	&	87.4	/	90.7	&	88.7	&	90.7	&	77.9	&	87.5	&	89.1	/	84.8	&	70	&	84.5	\\	
																														
	&		&	16	&	80.8	/	81.4	&	86.9	/	90.2	&	89.3	&	91.5	&	79.2	&	87.7	&	90	/	85.8	&	69.7	&	84.8	\\	
	&	SM	&	32	&	80.7	/	81.4	&	86.9	/	90.3	&	88.7	&	90.6	&	79.3	&	87.7	&	90.4	/	86	&	70.4	&	84.8	\\	
	&		&	64	&	80.7	/	81	&	87.1	/	90.5	&	89.1	&	90.9	&	79	&	88.1	&	90.4	/	86.3	&	72.2	&	\textbf{85.0}	\\	\hline
																														
	&		&	16	&	81.2	/	81.5	&	90.3	/	86.8	&	89.0	&	90.9	&	78.7	&	87.5	&	90.2	/	86.3	&	69.7	&	\textbf{84.7}	\\	
	&	SM $cf$=$0.5$	&	32	&	80.8	/	81.1	&	90.4	/	86.9	&	89.0	&	90.4	&	78.6	&	87.9	&	90.8	/	86.8	&	68.6	&	\textbf{84.7}	\\	
	&	\ \  (FSM)	&	64	&	80.6	/	81.2	&	90.5	/	87.3	&	88.5	&	90.9	&	78.8	&	88.7	&	89.6	/	85	&	69	&	84.6	\\	
																														
	&		&	16	&	80.8	/	81.7	&	86.6	/	90.2	&	88.8	&	90.8	&	78.3	&	88	&	89.1	/	84.3	&	70.4	&	\textbf{84.5}	\\	
	&	SM $g$=$2048$	&	32	&	80.5	/	81.6	&	86.6	/	90.3	&	88.5	&	90.5	&	77.7	&	87.9	&	89.4	/	85	&	69.3	&	84.3	\\	
\ref{tab:faster_sparse_mixers}	&		&	64	&	81.2	/	81.7	&	87.1	/	90.4	&	88.6	&	90.5	&	78.2	&	88.7	&	88.8	/	84.1	&	68.6	&	84.4	\\	
																														
	&		&	16	&	80.1	/	80.7	&	86.8	/	90.2	&	88.3	&	91.7	&	76.4	&	86.9	&	88.4	/	83.3	&	68.6	&	83.8	\\	
	&	SM $g$=$1024$	&	32	&	80.7	/	81	&	86.9	/	90.3	&	88.2	&	90.6	&	77.3	&	87.3	&	88.7	/	83.6	&	68.6	&	83.9	\\	
	&		&	64	&	80.3	/	80.9	&	90.4	/	86.9	&	88.2	&	90.5	&	78.2	&	87.7	&	88.5	/	83.6	&	70.4	&	\textbf{84.1}	\\	
																														
	&		&	16	&	80.1	/	81.1	&	86.8	/	90.4	&	88.4	&	90.9	&	76.4	&	87.7	&	88.6	/	83.8	&	67.9	&	83.8	\\	
	&	SM $cf$=0.75,	&	32	&	80.6	/	80.9	&	87	/	90.4	&	88.2	&	90.8	&	76.6	&	88.5	&	88.3	/	83.6	&	68.6	&	84.0	\\	
	& \quad \ $g$=$2048$	&	64	&	80.1	/	81.3	&	87.3	/	90.6	&	88.4	&	91.3	&	78	&	88.2	&	89	/	84.3	&	69	&	\textbf{84.3}	\\	\hline
																														
	&		&	16	&	80.7	/	81.2	&	90.5	/	87	&	88.2	&	90.8	&	78.1	&	87	&	88.7	/	83.8	&	69	&	\textbf{84.1}	\\	
\ref{tab:stability}	&	BERT-12	&	32	&	80.9	/	81.3	&	87.1	/	90.5	&	88.2	&	91.2	&	78.4	&	87.1	&	88	/	82.8	&	65.3	&	83.7	\\	
	&		&	64	&	80.9	/	80.7	&	86.5	/	90.1	&	88.5	&	90.9	&	79.1	&	87.2	&	87.4	/	81.8	&	65.7	&	83.5	\\	\hline
         
    \end{tabular}
\end{table*}

\begin{table*}
    \caption{SuperGLUE results (\emph{Validation} split) for final comparison of BERT, Sparse Mixer (SM), Fast Sparse Mixer (FSM) and other variants for different batch sizes. See the corresponding table for descriptions of each configuration. We report macro-F1 scores for CB, micro-F1/exact match scores for MultiRC, F1/exact match scores for ReCoRD, and accuracy scores for all other tasks. For each task, we select the best result across the base learning rates $\{10^{-5}, 5 \cdot 10^{-5}, 10^{-4}, 5 \cdot 10^{-4}, 10^{-3}\}$. The highest average score for each model (across the three batch sizes) is highlighted in boldface.}
    \label{tab:super_glue_batch}
    \small
    \centering
    \begin{tabular}{c | c | c | c c c c c c c | c}
        \hline
         Table & Model & Batch  & BoolQ & CB & COPA & MultiRC & ReCoRD & RTE & WiC & Avg. \\ \hline \hline
         
         	&		&	16	&	74.6	&	86.4	/	85.7	&	58	&	74.1	/	26.2	&	68.6	/	52.2	&	65	&	65.7	&			\textbf{65.7}	\\	
	&	BERT	&	32	&	72.9	&	85	/	85.7	&	61	&	73.7	/	25.6	&	70.4	/	54.4	&	58.5	&	67.9	&			65.5	\\	
\ref{tab:super_glue}	&		&	64	&	71.3	&	75.2	/	80.4	&	66	&	71.7	/	21.3	&	69	/	52.7	&	62.5	&	67.9	&			63.8	\\	
																														
	&		&	16	&	74.4	&	93.3	/	92.9	&	62	&	72.4	/	22.5	&	65.9	/	49.2	&	64.6	&	66.5	&			\textbf{66.4}	\\	
	&	SM	&	32	&	73.4	&	86.5	/	87.5	&	62	&	72.8	/	23.4	&	66.3	/	49.6	&	67.9	&	66	&			65.5	\\	
	&		&	64	&	72.4	&	89.7	/	89.3	&	59	&	72.6	/	23.3	&	65.7	/	48.9	&	68.6	&	61.8	&			65.1	\\	\hline
																														
	&		&	16	&	73.5	&	85.2	/	85.7	&	69	&	71.6	/	21.1	&	65.6	/	48.8	&	67.9	&	67.7	&			\textbf{65.6}	\\	
	&	FSM	&	32	&	74.3	&	82.3	/	85.7	&	56	&	71.8	/	21.5	&	63.8	/	46.9	&	67.1	&	62.4	&			63.2	\\	
	&  (SM w/ $cf$=$0.5$)	&	64	&	73.2	&	87.8	/	87.5	&	58	&	70.1	/	19.9	&	65.3	/	48.5	&	68.2	&	62.7	&			64.1	\\	
																														
	&		&	16	&	74.4	&	85.7	/	86.1	&	64	&	70.6	/	20	&	63.2	/	46.2	&	66.8	&	61.4	&			63.8	\\	
	&	SM w/ $g$=$2048$	&	32	&	74.1	&	82.8	/	83.9	&	63	&	70.7	/	20.9	&	63.6	/	46.7	&	66.8	&	62.2	&			63.5	\\	
\ref{tab:faster_sparse_mixers}	&		&	64	&	74.2	&	85.2	/	89.3	&	64	&	71.1	/	20.8	&	66.6	/	49.9	&	69	&	60.7	&			\textbf{65.1}	\\	
																														
	&		&	16	&	73.9	&	78.7	/	80.4	&	63	&	70	/	18.7	&	62.8	/	45.8	&	64.3	&	63.5	&			62.1	\\	
	&	SM w/ $g$=$1024$	&	32	&	73.4	&	86.6	/	85.7	&	64	&	70.1	/	18.4	&	64.5	/	47.6	&	66.4	&	61.4	&			63.8	\\	
	&		&	64	&	73.6	&	86.9	/	85.7	&	61	&	69.2	/	19.4	&	64.8	/	47.9	&	68.6	&	63	&			\textbf{64.0}	\\	
																														
	&		&	16	&	73.1	&	92	/	89.3	&	67	&	69.5	/	19.9	&	62.4	/	45.4	&	67.1	&	66.1	&			\textbf{65.2}	\\	
	&	SM w/ $cf$=0.75,	&	32	&	73	&	87.7	/	89.3	&	60	&	69.6	/	19.9	&	63.9	/	46.9	&	66.4	&	64	&			64.1	\\	
	&	\qquad\quad $g$=$2048$	&	64	&	72.4	&	80.1	/	82.1	&	62	&	69	/	18.9	&	65	/	48.1	&	67.1	&	63.3	&			62.8	\\	\hline
																														
	&		&	16	&	73.9	&	90.5	/	92.9	&	61	&	65.3	/	9.5	&	48.5	/	32	&	66.8	&	59.6	&			60.0	\\	
\ref{tab:stability}	&	BERT-12	&	32	&	72.5	&	92.3	/	92.9	&	58	&	69.6	/	15.5	&	50	/	33.3	&	65	&	60.2	&			\textbf{60.9}	\\	
	&		&	64	&	70.1	&	87.8	/	91.1	&	66	&	64.9	/	6.8	&	45.4	/	29.5	&	65.7	&	61	&			58.8	\\	\hline
         
    \end{tabular}
\end{table*}

\subsection{Full GLUE and SuperGLUE results}
\label{subsec:full_glue}

Table \ref{tab:glue_coord} contains the full GLUE results for all of the coordinate descent experiments summarized in Section \ref{sec:coord_descent}. Tables \ref{tab:glue_batch} and \ref{tab:super_glue_batch} contain the full results for GLUE and SuperGLUE, respectively, across all fine-tuning batch sizes, for the final model results tabulated in Sections \ref{sec:sparse_mixer}.

\subsection{Optimizing fine-tuning protocols for MoE models}
\label{subsec:moe_finetuning}

In Table \ref{tab:moe_finetuning_protocols}, we compare GLUE results for different fine-tuning learning protocols. Consistent with \cite{zoph2022designing}, we find that fine-tuning results are improved with an increased expert dropout (0.2) and larger base learning rates. Increasing the capacity factor, $cf$, yields quality gains but slows down the model. We did not find benefits from freezing different parts of the model during fine-tuning. 

For MoE models in the main text, we always use an expert dropout of 0.2 during fine-tuning. Similarly, we use the larger base learning rates during the coordinate descent program (Section \ref{sec:coord_descent}). However, when comparing Sparse Mixer with BERT (Section \ref{sec:sparse_mixer}) we use a wider range of learning rates for both models.

\begin{table*}
    \caption{Optimizing fine-tuning learning protocols for MoE models. The "Default LR" range is $\{10^{-5}, 5 \cdot 10^{-4}, 10^{-4}\}$, while the "Large LR" range is $\{10^{-4}, 5 \cdot 10^{-4}, 10^{-3}\}$. For "Fine-tune All", we fine-tune all layers; for "Freeze MoE", we fine-tune all but the MoE layers; and for "Dense MLP", we only fine-tune the dense MLP sublayers. The default model is repeated in several rows: "0.2 ex dropout", "Default LR" (learning rate), "$cf$=1.", and "Fine-tune All". The top average scores are highlighted in boldface.}
    \label{tab:moe_finetuning_protocols}
    \centering
    % \small
    \setlength{\tabcolsep}{5pt}
    \begin{tabular}{l | c c c c c c c c | c}
        \hline
         Configuration & MNLI & QQP & QNLI & SST-2 & CoLA & STS-B & MRPC & RTE & Avg. \\ \hline \hline
         0.0 ex dropout	&	80	/	80.8	&	87	/	90.4	&	88	&	89.9	&	75.7	&	87.1	&	86.1	/	79.9	&	66.1	&	82.8	\\	
0.1 ex dropout	&	80.2	/	81.3	&	87.1	/	90.3	&	87.7	&	90	&	75.8	&	87	&	86.3	/	80.4	&	65.7	&	82.9	\\	
0.2 ex dropout	&	80.7	/	81.3	&	86.9	/	90.3	&	87.9	&	90.6	&	76.8	&	86.8	&	86.1	/	79.7	&	66.8	&	\textbf{83.1}	\\	
0.3 ex dropout	&	80.5	/	81.6	&	87	/	90.3	&	87.9	&	90.4	&	76.2	&	86.7	&	86.1	/	80.1	&	66.1	&	83.0	\\	
0.4 ex dropout	&	80.6	/	81.5	&	86.8	/	90.3	&	88.2	&	90.5	&	76	&	87.5	&	86.5	/	80.4	&	65.3	&	\textbf{83.1}	\\	\hline

\multicolumn{10}{}{} \\ \hline
Default	LR &	80.7	/	81.3	&	86.9	/	90.3	&	87.9	&	90.6	&	76.8	&	86.8	&	86.1	/	79.7	&	66.8	&	83.1	\\
Large LR	&	80.4	/	81.1	&	86.8	/	90.2	&	87.5	&	90.6	&	75.9	&	84.9	&	88.7	/	83.6	&	68.6	&	\textbf{83.5}	\\	\hline

\multicolumn{10}{}{} \\ \hline
$cf$=1. &	80.7	/	81.3	&	86.9	/	90.3	&	87.9	&	90.6	&	76.8	&	86.8	&	86.1	/	79.7	&	66.8	&	83.1	\\
$cf$=2.	&	80.1	/	81.1	&	86.8	/	90.2	&	88	&	90.7	&	77.4	&	86.9	&	87	/	82.1	&	66.4	&	\textbf{83.3}	\\	\hline
							
\multicolumn{10}{}{} \\ \hline	
Fine-tune All &	80.7	/	81.3	&	86.9	/	90.3	&	87.9	&	90.6	&	76.8	&	86.8	&	86.1	/	79.7	&	66.8	&	\textbf{83.1}	\\
Freeze MoE	&	80.4	/	80.9	&	86.9	/	90.2	&	88.2	&	90.9	&	75.3	&	86.6	&	85.9	/	78.9	&	67.1	&	82.8	\\	
Dense MLP	&	78.8	/	79	&	89.9	/	86.4	&	87.3	&	90.3	&	69.2	&	84.4	&	73.5	/	83.2	&	63.2	&	80.5	\\	\hline
    \end{tabular} 
\end{table*}

\subsection{Speed-accuracy plots}
\label{subsec:speed_nsp_accuracy}

Figure \ref{fig:nsp_scale} shows the NSP-accuracy equivalent of the MLM-accuracy based efficiency plot in Figure \ref{fig:mlm_scale}. 

\begin{table*}
    \caption{Model configurations. Following the Base convention for both BERT and Sparse Mixer, we set $d_{ff}=4d_{h}$ and the number of self-attention heads to $d_{h}/64$. Roughly following the Sparse Mixer Base design, we set the number of MoE and attention layers to be roughly 25-33\% of the larger models and no less than 2 for the smaller models. We increase the number of experts for the larger models.}
    \label{tab:scale_sizes}
    \centering
    \begin{tabular}{l | c c c | c c c c c c}
        \hline
        & \multicolumn{3}{c|}{BERT} & \multicolumn{6}{c}{Sparse Mixer}  \\ 
        Name & Layers & $d_{m}$ & $d_{ff}$ & Layers & $d_{m}$ & $d_{ff}$ & Attention & MoE & Experts \\ \hline \hline
        & 2 & 256 & 1024 & 2 & 256 & 1024 & 2 & 2 & 16 \\
        & 4 & 256 & 1024 & 4 & 256 & 1024 & 2 & 2 & 16 \\
        & 4 & 512 & 2048 & 4 & 512 & 2048 & 2 & 2 & 16 \\
        & 8 & 512 & 2048 & 8 & 512 & 2048 & 4 & 4 & 16 \\
        Base & 12 & 768 & 3072 & 14 & 512 & 2048 & 4 & 4 & 16 \\
        & 18 & 768 & 3072 & 18 & 768 & 3072 & 6 & 6 & 32 \\
        Large & 24 & 1024 & 4096 & 24 & 1024 & 4096 & 6 & 6 & 64 \\ \hline
    \end{tabular}
\end{table*}

Table \ref{tab:scale_sizes} gives the model configurations that were used to construct Figure \ref{fig:mlm_scale} (main text) and Figure \ref{fig:nsp_scale} (Appendix \ref{subsec:speed_nsp_accuracy}). In configuring the Sparse Mixer model configurations, we tried to roughly hew to the proportions in the Base model.

\end{document}